\documentclass[preprint,12pt,authoryear]{elsarticle}

\usepackage{natbib}
\usepackage{booktabs}
\usepackage{adjustbox} 
\usepackage{caption}
\usepackage{subcaption}

\usepackage[dvipsnames,table]{xcolor}

\usepackage[nomain, nonumberlist, acronym, shortcuts]{glossaries}
\makeglossaries
\glsenablehyper
\newacronym{ai}{AI}{Artificial Intelligence}
\newacronym{ann}{ANN}{Artificial Neural Networks}
\newacronym{anova}{ANOVA}{Analysis of Variance}
\newacronym{ap}{AP}{Average Precision}
\newacronym{ar}{AR}{Average Recall}
\newacronym[plural={BBoxes}, \glsshortpluralkey={BBoxes}]{bbox}{BBox}{Bounding Box}
\newacronym{cau}{CAU}{Christian-Albrechts-Universität }
\newacronym{cnn}{CNN}{Convolutional Neural Network}
\newacronym{coco}{COCO}{Common Objects in Context}
\newacronym{csp}{CSP}{Cross-Stage-Partial-connections}
\newacronym{dab-detr}{DAB-DETR}{Dynamic Anchor Box DETR}
\newacronym{detr}{DETR}{DEtection TRansformer}
\newacronym{dl}{DL}{Deep Learning}
\newacronym{dnn}{DNN}{Deep Neural Network}
\newacronym{fcn}{FCN}{Fully Connected Layer}
\newacronym{fn}{FN}{False Negativ}
\newacronym{fp}{FP}{False Positiv}
\newacronym{gpu}{GPU}{Graphics Processing Unit}
\newacronym{gt}{GT}{Ground Truth}
\newacronym{hpc}{HPC}{Hyperparameter Combination}
\newacronym{iou}{IoU}{Intersection over Union}
\newacronym{ktbl}{KTBL}{Kuratorium für Technik und Bauwesen in der Landwirtschaft e.V.}
\newacronym{mig}{MIG}{Multi-Instance GPU}
\newacronym{ml}{ML}{Machine Learning}
\newacronym{resnet}{ResNet}{Residual Network}
\newacronym{rl}{RL}{Reinforcement Learning}
\newacronym{rpn}{RPN}{Region Proposal Network}
\newacronym{r-cnn}{R-CNN}{Region-based Convolutional Neural Network}
\newacronym{roi}{RoI}{Region of Interest Pooling}
\newacronym{sl}{SL}{Supervised Leraning}
\newacronym{tp}{TP}{True Positiv}
\newacronym{ul}{UL}{Unsupervised Learning}
\newacronym{yolo}{YOLO}{You Only Look Once}

\glsaddall

\usepackage{amssymb}
\usepackage{amsmath}
\usepackage{lineno}

\usepackage[pdftex, colorlinks=false, pdfborder={0 0 0}]{hyperref}
\usepackage{orcidlink}

\journal{Computers and Electronics in Agriculture}

\begin{document}

\begin{frontmatter}



\title{Excretion Detection in Pigsties Using Convolutional and Transformer-based Deep Neural Networks} %% Article title


\author[aff1,cor1]{Simon Mielke\orcidlink{0009-0008-6490-2720}} %% Author name
\author[aff1,cor1]{Anthony Stein\orcidlink{0000-0002-1808-9758}} %% Author name

\affiliation[aff1]{organization={University of Hohenheim, Department of Artificial Intelligence in Agricultural Engineering},%Department and Organization
            addressline={Garbenstraße 9}, 
            city={Stuttgart},
            postcode={70599}, 
            state={Germany},
            country={Baden-Wuerttemberg}}

\cortext[cor1]{Correspondence to: Simon Mielke, \href{mailto:simon.mielke@uni-hohenheim.de}{simon.mielke@uni-hohenheim.de} and Anthony Stein, \href{mailto:anthony.stein@uni-hohenheim.de}{anthony.stein@uni-hohenheim.de}}

\newpageafter{author}
%% Abstract
\begin{abstract}
Animal excretions in form of urine puddles and feces are a significant source of emissions in livestock farming. 
Automated detection of soiled floor in barns can contribute to improved management processes but also the derived information can be used to model emission dynamics. 
Previous research approaches to determine the puddle area require manual detection of the puddle in the barn. 
While humans can detect animal excretions on thermal images of a livestock barn, automated approaches using thresholds fail due to other objects of the same temperature, such as the animals themselves. 
In addition, various parameters such as the type of housing, animal species, age, sex, weather and unknown factors can influence the type and shape of excretions. 
Due to this heterogeneity, a method for automated detection of excretions must therefore be not only be accurate but also robust to varying conditions. 
These requirements can be met by using contemporary deep learning models from the field of artificial intelligence. 
In particular, machine vision models such as object detection and segmentation through supervised learning offer a promising solution to deal with this challenge. 
This work is the first to investigate the suitability of different deep learning models for the detection of excretions in pigsties, thereby comparing established convolutional architectures with recent transformer-based approaches. 
%As an exemplary case, two training data sets from two pig houses are created and annotated. 
The state-of-the-art object detection models Faster R-CNN, YOLOv8, DETR and DAB-DETR are compared and statistically assessed on two created training datasets representing two pig houses. 
%The models are preconfigured from MMDetection's and MMYOLO's model Zoo for the COCO dataset. 
%Three hyperparameters (batch size, class loss weighting and training status) are varied and tested in different combinations. 
To this end, we apply a method derived from nested cross-validation and report on the results in terms of eight common detection metrics. 
Our work demonstrates that all investigated deep learning models are generally suitable for reliably detecting excretions with an average precision of over 90\%. 
The models also show robustness on out of distribution data that possesses differences from the conditions in the training data, however, with expected slight decreases in the overall detection performance. 
We discuss the observed significant differences between the compared models and elaborate on important insights regarding the choice of the models' hyperparameters as well as the data requirements. 
%Further research is needed, especially on the configuration of the hyperparameters.

\end{abstract}

%\includegraphics[width=\textwidth]{Graph-Abstract}

%\input{highlights.tex}

%% Keywords

\end{frontmatter}

%\linenumbers



\section{Introduction}
\label{sec:Introduction}
\noindent
Emissions from agriculture accounted for around 8\% of total emissions in Germany in 2023 and approximately 68.1\% of this is attributable to livestock farming \citep{UBA_THG}. The proportion of ammonia emissions in Germany attributable to agriculture is around 95\% and the proportion attributable to livestock farming is 70\% \citep{UBA_AGS}. For nitrous oxide (N$_{2}$O), the proportion of emissions from agriculture in 2022 was around 67\% and for methane (CH$_{4}$)  76\% of total nitrous oxide and methane emissions in Germany\citep{UBA_LuM}. Agriculture is therefore a relevant source of emissions and greenhouse gases in Germany \citep{UBA_THG} and also worldwide \citep{global_greenhouse-gas_emissions}. Research has already been carried out into methods of measuring and reducing emissions from dairy cattle and fattening pig barns and emission data has been collected \citep{EmiDat,EmiDat_2,EmiMin,cow_atmospheric_flux}. This made it possible to validate and establish measurement standards, create a uniform data basis for emissions and test reduction methods.
\\
So far a basic requirement for measuring the emission area on the floor is always the prior manual detection of the animal excretions by a person. Depending on the method, the resulting limitations or inaccuracies include the influence of animal behavior by the observers, the subjective detection and marking of the animal excretions as well as the personnel and time required or the limitation in the time of day which can be investigated. The parallel detection of several excretions also represents a challenge.
\\
\ac{ai} models for recognizing objects can be found in the literature targeting the agricultural sector of livestock farming. For instance, the detection of the position and posture of pigs is an important indicator for assessing the behavior of the animals and their well-being. Such behavior monitoring tasks have been automated using an AI approach to object detection \citep{position-posture_deep-learning,automated_pig_behavior,behaviour_pig_devormable-concolution}. Other approaches to object detection are dealing with the parallel identification and tracking of several pigs or cows during the day and at night \citep{automatic_individual_tracking,individual_pig_detection_tracking,tracking_online_dense-groups,cow-multi_tracking}.
\\
For the detection of water puddles, various computer vision methods have been investigated in the past in addition to other methods \citep{water_20-year_review}. In particular, the detection of water puddles on the roadway of cars, unmanned vehicles and mobile robots with \ac{ann}s and \ac{cnn}s was investigated. This was based on the reflective property of water puddles, laser reflections or a stereo camera and in some cases a corresponding dataset was created \citep{water_FCN_refelction-attention,water_puddle_laser-reflection,puddle_all_CNN_real-time,water_puddle_stereo-camera}. However, the characteristics (shape, reflection, environment, formation) and the data basis (RGB/IR images) differed between water puddles and the urine puddles described in \autoref{sec:Background}.
\\
This work is the first to investigate the suitability of AI-based computer vision models for the detection of animal excretions using thermal infrared images. While this is possible for humans, simpler automated approaches using thresholds fail to detect the excretions. This however results in a high manual workload and inaccuracies due to the subjective detection by the human observer. Both continual detection of excretions and area-wide detection of all excretions in barns are thus barely possible. In addition, many different and variable parameters such as housing type, animal species, animal age, sex, weather and possibly unknown parameters can have an effect on the type and shape of the excretions. The method required for detection must therefore be robust and adaptive to a high variety of local and environmental conditions. These requirements in turn call for the strengths of \ac{ml} models from the field of \ac{ai}. Models for computer vision, or more precisely object detection or segmentation from the field of \ac{sl}, constitute promising approaches that have as of yet not been investigated with regard to the conditions and the task at hand.
\\
This work therefore proposes a novel approach for the basic identification and detection of animal excretions in pig pens using deep learning-based computer vision methods. Object detection specifically for infrared images of urine puddles and feces will be tested with various architectures already established on other large datasets, such as Faster \ac{r-cnn} \citep{Software_Faster_R-CNN} and \ac{yolo}v8 \citep{Software_YOLOv8} using transfer learning. Furthermore, vision transformer-based detection models such as \ac{detr} and \ac{dab-detr} will be tested \citep{Software_DETR,Software_DAB-DETR}. The creation and thorough annotation of a corresponding and representative training dataset has also been part of the work. This special dataset with thermal images from pigsties and the annotation of the excretions as object class are novel features of this work.
\\
The objectives of the work are 1) to investigate whether our assumption of the suitability of deep learning-based computer vision models for the detection of animal excretions holds true, 2) to compare the performance of different models in terms of several common metrics and 3) to create an annotated animal excretion dataset comprised of thermal infrared images. Accordingly, the overarching contribution of this work is the identification and discussion of the model and data requirements that pave the way to achieve an accurate detection of animal excretions. For this purpose, hyperparameters considered especially relevant for this task are systematically varied. Finally, the robustness of the models will be tested and compared under different conditions, such as different locations, housing types or light and temperature conditions.

\section{Background}
\label{sec:Background}
\begin{figure}[h]
  \centering
  \begin{subfigure}{0.49\linewidth}
    \centering
    \includegraphics[width=\linewidth]{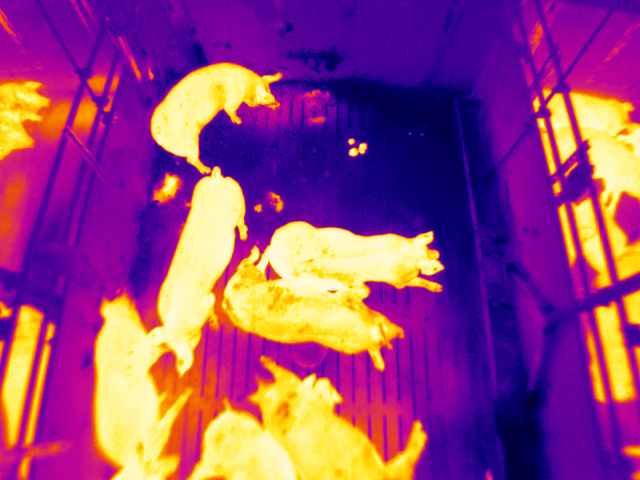}
    \caption{Original thermal image. Urine puddles and pigs are clearly distinguishable from the floor and from each other.}
    \label{fig:Schwellenwert0}
  \end{subfigure}
  \hfill
  \begin{subfigure}{0.49\linewidth}
    \centering
    \includegraphics[width=\linewidth]{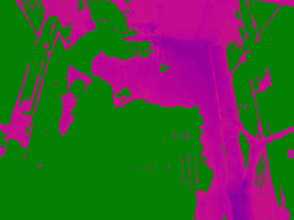}
    \caption{Image after applying a threshold value. Pigs and large areas of the heated floor are recognized as puddles.}
    \label{fig:Schwellenwert2}
  \end{subfigure}
  \caption[Detection of urine puddles with threshold value method]{Detectability of urine puddles after application of a threshold value method over the entire recording area.}
  \label{fig:Schwellenwert}
\end{figure}

\noindent
Nitrogen is excreted with the urine of pigs in the form of urea (CH$_{4}$N$_{2}$O), which is previously formed in the liver during the detoxification of ammonia and then transported to the kidneys via the blood. To a certain extent, this process cannot be avoided due to the metabolism of proteins and amino acids in the body \citep{UTB-Anatomie}. Growing livestock in particular, such as fattening pigs, have a constantly changing need for protein and the amino acids required for this \citep{UTB-Nahrung}. Depending on the housing system, management and requirements of various institutions for agriculture, pigs may be oversupplied with amino acids. Such an oversupply is almost completely reflected in an increased excretion of urea via the urine. The requirements of organic farming, for example, include a ban on the use of free or synthetic amino acids \citep{EU}. The consequences are a worse adaptation of the supply of amino acids to the needs of the animals and the associated oversupply and increased excretion of urea via the urine \citep{UTB-Nahrung}. Urease is an enzyme and is found in large parts of the environment, as it can be produced by various bacteria, yeasts, fungi and algae \citep{Urease_microbial}. Microorganisms in the forestomachs of ruminants synthesize urease to utilize recycled urea \citep{UTB-Anatomie}. In the digestive tract of monogastric animals, especially in the colon, microorganisms also produce urease to break down endogenous urea \citep{Urease_intestinal-anaerobes,UTB-Nahrung}. The microorganisms from the forestomachs and colon are excreted with the feces. The activity of urease in fresh pig feces \citep{Urease_activity_ammonia-production} is correspondingly high. As soon as urease encounters urea in an aqueous environment, the enzyme catalyses the hydrolysis of the urea molecules to ammonia (NH$_{3}$) and carbon dioxide (CO$_{2}$) \citep{Urease_microbial, Urea_urease}. The volatilization of the resulting ammonia follows Henry's law and is therefore proportional to the ammonia concentration in the liquid or urine puddle. The activity of the urease is related to the temperature \citep{Urease_activity_ammonia-production}. The hydrolysis of urea to ammonia is in turn subject to the concentration of urease and microorganisms. The volatilization of the ammonia depends on the temperature, the air velocity at the boundary layer between the air and the puddle and the surface of the puddle \citep{ammonia_infuence-factors_migration_techniques}. The contact of feces and urine on the same surface therefore significantly influences the formation of ammonia, which is why puddles of urine especially in combination with feces are a significant source of emissions in livestock farming. 
\\
Ammonia is a reactive gas that is known for its much-discussed negative environmental effects and is referred to as an indirect greenhouse gas. Its further conversion produces nitrous oxide, which is classified as around 265 times more harmful to the climate than carbon dioxide \citep{ammonia_atmospheric_effects,UBA_LuM}. In addition to ammonia, mixing feces and urine also produces emissions such as odor, particulate matter, bioaerosols and, as already mentioned, carbon dioxide \citep{UBA_AGS}. Another aspect that should not be neglected is the economic perspective, which focuses on the loss of nitrogen from the manure and the subsequent reduced effectiveness of the manure as a fertilizer.
\\
Several approaches to estimating and measuring emissions from urine puddles have already been researched and various models developed \citep{ammonia_modelling_naturally_part1,ammonia_dynamic_model_slatted-floors,ammonia_dynamic_model,ammonia_computer-modelling_emission-rates,ammonia_group-housing_straw-bedding,cow_NH3_prediction_reaction_kinetic_modeling}. The area of the puddles is one of the five most important variables for modeling the resulting emissions \citep{cow_mechanistic_models_NH3_puddle}. Determining the emission area of urine puddles in turn requires detecting the puddles and their boundaries. As a consequence the ability to model and estimate emissions in agricultural livestock farming on the one hand and to reduce them efficiently and in a targeted manner on the other hand requires, among other things, the identification and detection of urine puddles and feces.
\\
Urine puddles in livestock farming have a number of characteristics that distinguish them from conventional water puddles. As a rule, large puddles do not accumulate on the floors of barns, as large quantities of liquids normally run off. Rather than being puddles of water with urine instead of water, urine puddles, especially on slatted floors, are simply moist surfaces wet with urine that are left behind. In addition, the shape of these surfaces can sometimes have clear and straight edges due to the slats on the floor. Therefore wetted surfaces are very difficult to detect in the spectrum of visible light. Visual methods to measure and determine the area of urine puddles or emission sources are mainly used by an individual person. These include, for example, measuring with a measuring frame or drawing a rectangle around the surface wetted with urine in the barn \citep{cow_double-sloped_frame,ammonia_effect_floor-fouling}. In tests under practical conditions, however, these methods proved to be inaccurate or unusable \citep{cow_IR-camera_urine-puddle}. In particular, it was not possible to detect the edge areas of fresh puddles, which are needed to distinguish them from other liquids and old puddles. The methods were therefore described as only feasible on dry soils \citep{cow_IR-camera_urine-puddle}. 
\\
On the other hand, the urine has at the time of excretion a temperature close to the animal's body temperature. This can be used as a key property for the detection of urine puddles with the help of infrared cameras and thermal imaging. An approach using a mobile infrared camera and subsequent differentiation between soil and urine puddles based on temperature thresholds exists \citep{cow_IR-camera_urine-puddle}. The definition of the threshold value was described as relevant for the resulting area. The higher the threshold temperature, the smaller the determined area. To solve this problem, a dynamic threshold method based on an average value or the subtraction of a background image after cooling the puddle was described \citep{cow_IR-camera_urine-puddle}. The method with a mobile infrared camera was further developed in a second approach with artificial, blue-colored puddles using the indicated dynamic threshold method \citep{cow_simulatet_puddle_IR}. Due to the blue color of the puddles, they could be seen and annotated in images in the visible range of light. The annotations were then used to validate the threshold method developed for the infrared images. The difficulties of the approach with a mobile camera were described as the manual detection of the puddles or urination, the time delay to get to the puddle with the camera and the undisturbed recording of the puddle \citep{cow_IR-camera_urine-puddle}. In addition, the possibility of the animals being influenced by the people carrying out the experiment cannot be ruled out. 
\\
A similar approach avoids this problem by using a stationary camera in the barn. While the mobile approach was developed for cattle, the stationary approach uses pigs as a livestock species with fixed places in the execution of their elimination behavior \citep{oberflaechenbenetzung_MA_CAU,pig_elimination_behavior}. The switch to a method with a stationary camera presented new challenges. In contrast to the mobile approach, in which only one puddle of urine was captured as the only object in the center of the images, the images now contain several puddles and various other warm objects can be seen next to the puddles. In addition, the puddle position is no longer in the center of the image. Despite the temperature as a key property, automated methods for detecting the urine puddles, such as threshold methods with fixed temperature ranges for the urine puddles, failed. The reason for this are other objects that have a similar or even the same temperatures as the puddles. The animals themselves and lying surfaces warmed up by them are examples of such objects. \autoref{fig:Schwellenwert} shows two images in the infrared range. One image before and one after applying a threshold method. This visualizes the basic problem of distinguishing puddles of urine from other objects of the same temperature and detecting puddles in different temperature ranges.

\noindent
The temperature profile of the puddles over time is also one of the properties relevant for detection. From the time of deposit, the puddles usually cool down within 10 minutes until they have reached the ambient temperature or the temperature of the ground \citep{cow_IR-camera_urine-puddle}. The cooling process can be used as an advantage for detection with the help of videos. However, the information of the temperature decrease over time is lost when working with images. The puddle has a different temperature depending on the time between the deposit of the puddle and the image being taken. Due to the decrease in temperature over time and the faster cooling of the urine puddles at the edges than in the center of the puddle, the area of the puddles determined with threshold values decreases over time. As no fixed temperature can be assigned to a puddle, the question arises as to the temperature, up to which a puddle can still be declared as such with fixed threshold value methods. 
\\
As a workaround for the detection and differentiation of the puddle area in the approach with the stationary camera, a digital measuring frame was therefore first manually drawn around the urine puddles in the images in order to separate the puddles from other objects. A threshold method was then used within the measurement frame to differentiate between the puddle and the ground \citep{oberflaechenbenetzung_MA_CAU}.
\\
Another factor that affects the visual detection of urine puddles are the different shapes and positions of the puddles between the sexes in relation to the animals. In females, the puddle is located behind the animal, whereas in males it is located under the animal. This makes it difficult to recognize the puddles from a 90 degree angle to the ground.
\\
Alongside swarm intelligence, evolutionary intelligence, problem-solving strategies and several other areas, \ac{ml} is part of the broad spectrum of the topics in \ac{ai}. The roots of \ac{ai} are widely scattered in mathematics, philosophy, economics, linguistics and the development of computers \citep{AI_modern-approach}. Applications with \ac{ai} are now an integral part of everyday life. Whether speech recognition, autonomous mowing robots, spam recognition for phone calls and emails or personalized recommendations for products or videos on various online platforms, some type of \ac{ai} is usually working in the background. The equally broad field of tasks and applications in agriculture offers a wide range of opportunities for the use and research of AI-supported solutions. 
\\
\ac{ml} itself is made up of several sub-disciplines. These include \ac{sl}, \ac{ul} and \ac{rl}.  What they all have in common is the fundamental principle of automatically adapting and improving the performance of machines or programs as a result of processing data. \ac{sl} uses annotated data in the form of input-output pairs. The aim is to learn a function that uses the input to generate the output \citep{AI_modern-approach}. This requires a representative and large dataset for training that provides as much data and as many types of input-output pairs as possible.  
\\
Computer vision is about processing visual data, such as images and videos. Therefor Methods such as \ac{sl} are often used. With \ac{sl} for object detection, the individual annotations are in the form of a rectangular \ac{bbox} with a label. The \ac{bbox} contains the information about the location (X and Y coordinates within an image) and size (height and width) of the object and the label or class of the object. During training, the object detection model learns a regression for the coordinates of the \acp{bbox} and a classification for the object it contains. Various model architectures exist to implement this. However, the precise and correct annotation of the data is always crucial for effective training of the models and often takes up most of the time within a project. Training a corresponding \ac{ai} model usually also requires a lot of time, computing power and large amounts of data. A common practice is therefore to train models on large, already annotated datasets in order to learn basic relationships. The ImageNet dataset or the \ac{coco} dataset are well-known examples of this \citep{Software_ImageNet,Software_COCO}. The pre-trained model is then transferred using transfer learning and trained on the smaller project-related datasets \cite{AI_transfer-learning_survey,AI_tranferable_features}.
\\
\Ac{ann} and \ac{dl} are central concepts of \ac{ml} in computer vision. A \ac{ann} consists of several layers with artificial neurons. An artificial neuron is a mathematical model that simulates the function of a real neuron in the brain. This refers to the reception, processing and transmission of stimuli and signals. Accordingly, an artificial neuron consists of an input function, an activation function and the resulting output. The original idea for this goes back to the year 1943 \citep{logical_calculus}. There are connections between the neurons of different layers, so that the output of one neuron or several neurons contributes to the input of one or more neurons of the next layer. These connections are weighted with parameters as factors. The \ac{ann} then learns by adjusting these parameters during training. 
\\
Since the layers and their results between the input and the output are only intermediate steps in the \ac{ann}, they are also called hidden layers. If a \ac{ann} has several hidden layers, it is referred to as a \ac{dnn} and \ac{dl}. Larger \ac{dnn}s with many layers have several million parameters and are therefore able to learn complex relationships. If there is a connection between every neuron of one layer to every neuron of the next layer, this is called a \ac{fcn}. Another type of \ac{ann} is a \ac{cnn}, where the weights are arranged in filters and thus neurons are only connected to certain neurons of other layers. In addition to the reduction of parameters, the use of filters enables the extraction and learning of features, which is why they are used in many computer vision models.

\section{Materials and Methods}
\label{sec:Materials and Methods}

\subsection{Data acquisition and preprocessing}
\begin{figure}[h!]
    \centering
    \begin{subfigure}{0.49\linewidth}
      \centering
      \includegraphics[width=\linewidth]{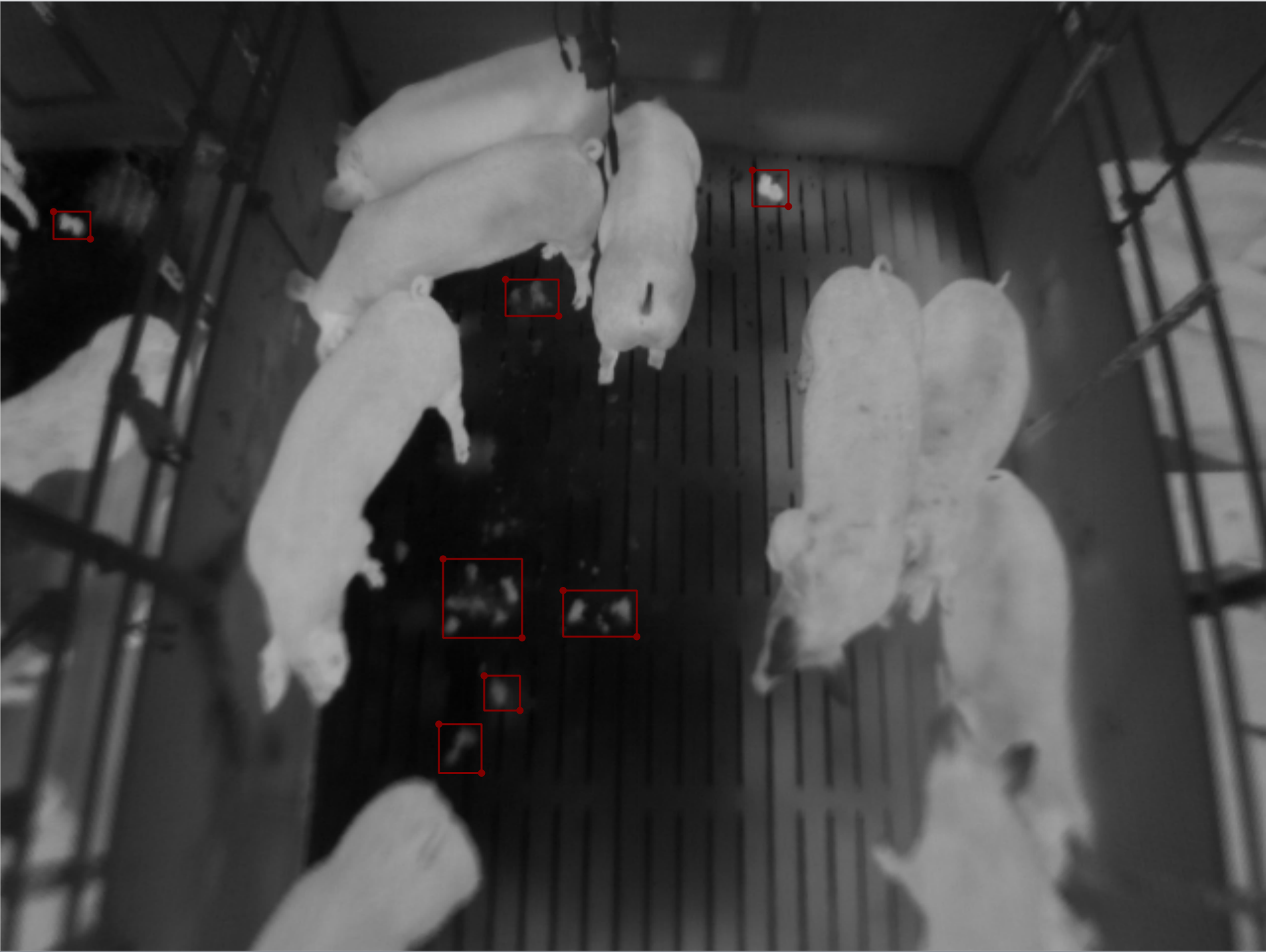}
      \caption{Annotated image from the first dataset with blurred areas outside the camera focus and urine puddles mixed with feces, making the differentiation unclear.}
      \label{fig:normal_vs_robust1}
    \end{subfigure}
    \hfill
    \begin{subfigure}{0.49\linewidth}
      \centering
      \includegraphics[width=\linewidth]{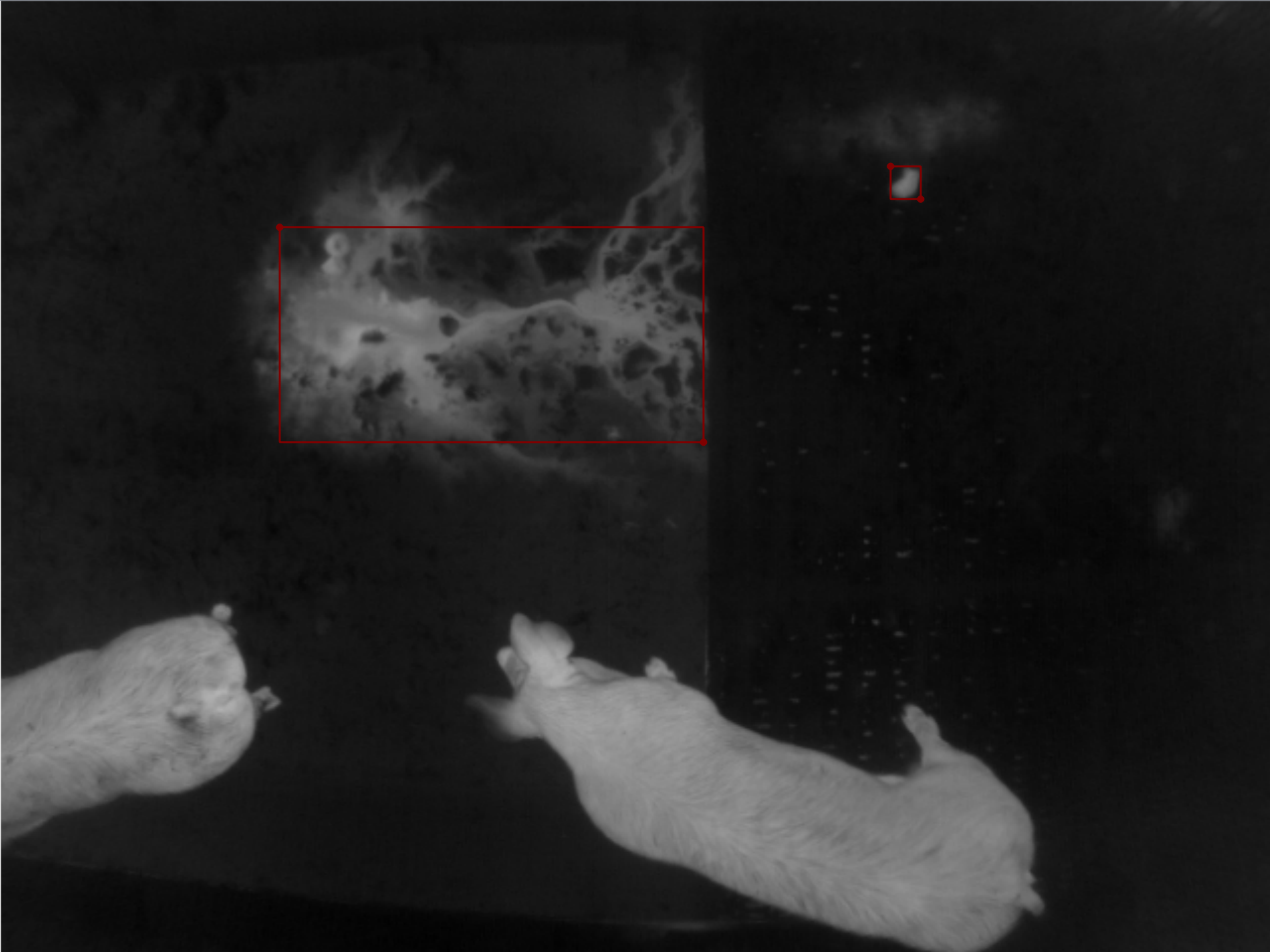}
      \caption{Annotated image from the second dataset with a different camera angle, a different distance of the camera to the ground, a different ground and the resulting larger puddle with a different structure.}
      \label{fig:normal_vs_robust2}
    \end{subfigure}
%%\end{figure}

%%\begin{figure}\ContinuedFloat
    \begin{subfigure}{0.49\linewidth}
      \centering
      \includegraphics[width=\linewidth]{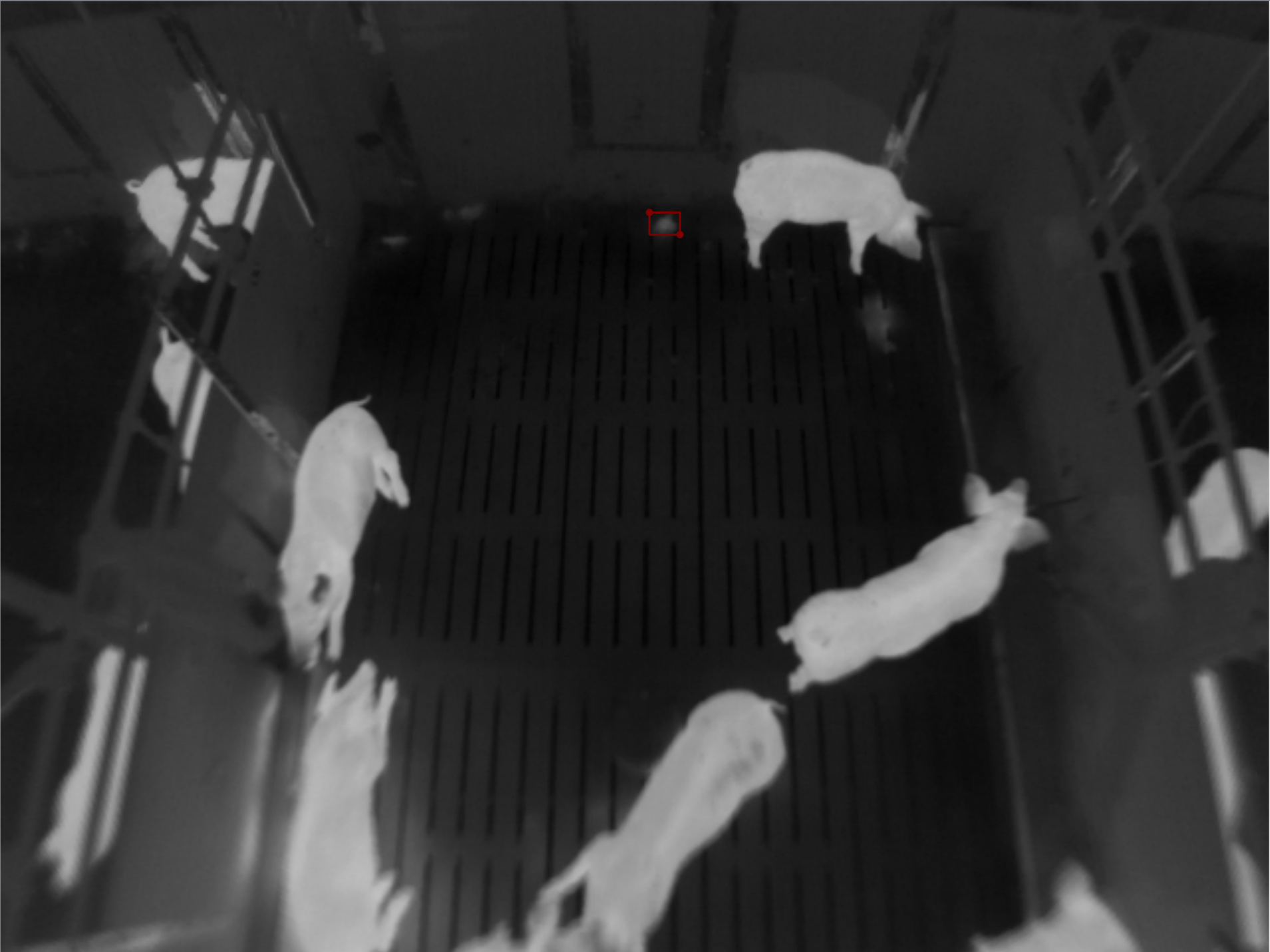}
      \caption{Annotated image of the first dataset with unclear object without temporal progression and video context under the top right pig.}
      \label{fig:normal_vs_robust3}
    \end{subfigure}
    \hfill
    \begin{subfigure}{0.49\linewidth}
      \centering
      \includegraphics[width=\linewidth]{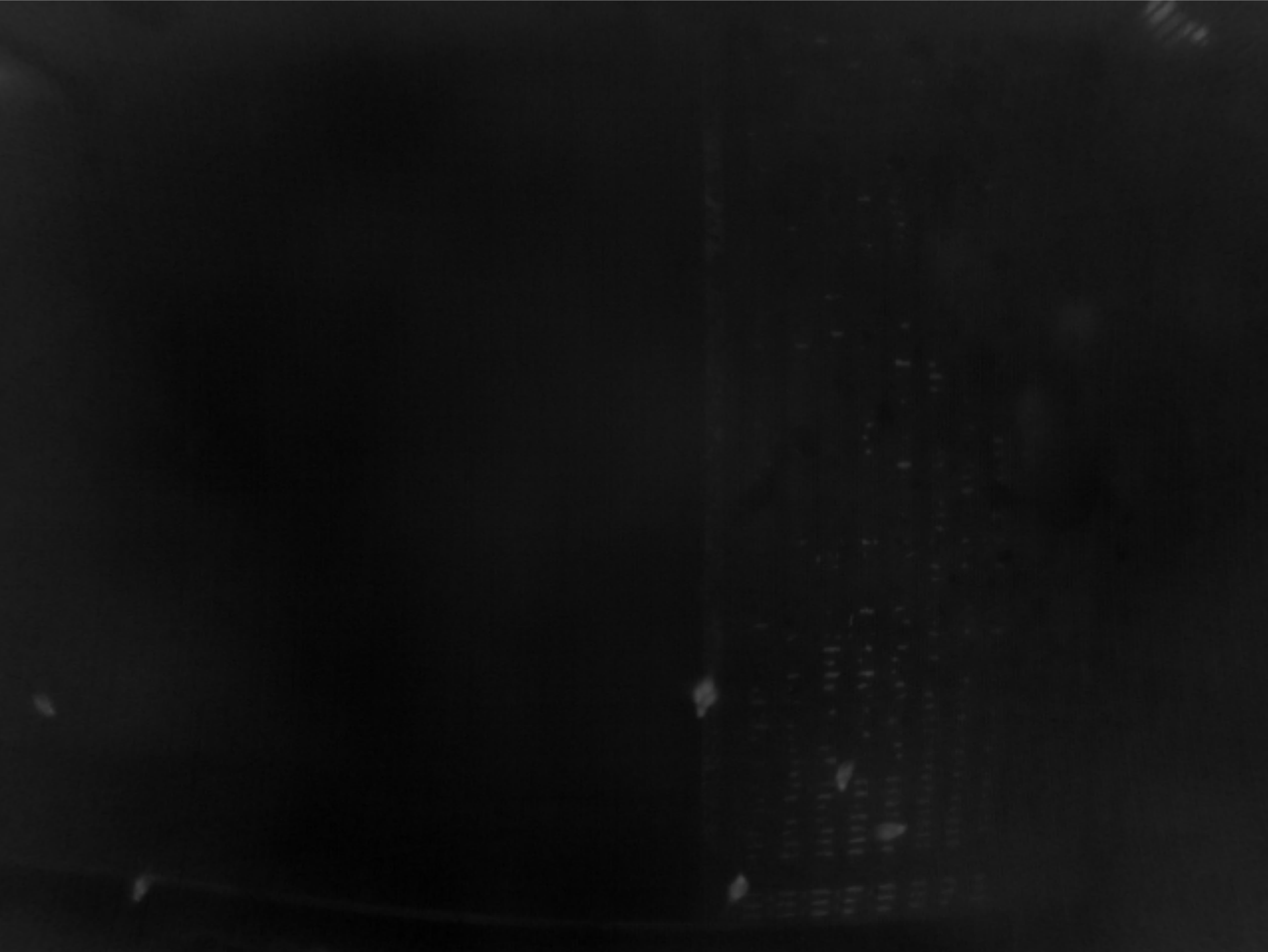}
      \caption{Image of the second dataset without excretions, but with birds and sun-warmed stripes on the ground in the upper right corner.}
      \label{fig:normal_vs_robust4}
    \end{subfigure}
    \caption[Annotation results]{Differences between datasets and challenges identified during annotation.}
    \label{fig:normal_vs_robust}
\end{figure}

\noindent
The first step of this work consists of creating an annotated dataset with thermal images for training the \ac{ai} models. A second test dataset with different properties is required to investigate the robustness of the trained  models. The first dataset, consisting of 1000 images, is created from images of a ventilated barn in Germany. For the second dataset, images from an freely ventilated barn in Germany with a different floor, a different camera angle and a different distance of the camera to the floor are processed. The sun could also partially shine in the freely ventilated barn. All images were taken from monochrome (grayscale) videos recorded with the optris PI 640 infrared camera from Optris GmbH (Berlin, Germany). The videos were recorded as part of the EmiMin project  prior to the work by \ac{cau} of Kiel, saved as a radiometric video file and made available for this work. The camera was stationary in a pen of the barns. Individual frames from the videos are then cut out as images. The resolution of the images and videos is $640*480$ pixels. As part of a master's thesis at \ac{cau}, 773 of the 1000 images of the first dataset have already been cut out with the optis PIX Connect software \citep{Software_PIX-Connect} and saved as false color images including raw data \citep{oberflaechenbenetzung_MA_CAU}. As part of the current work, further videos are viewed and 227 images were cut out with a script written in Python to expand the dataset.  The 200 images for the second test dataset from the freely ventilated barn are also based on individual frames of the video recordings, which are cut out during the process of this work.
\\
In order to obtain a uniform dataset, the processing of all images starts with the raw data. During the data preprocessing, all images are standardized to the same value range so that the same pixel values in different images also represent the same temperature. The raw data, which contains signed integers in 16-bit format was transferred in to grayscale images with unsigned integers in 8-bit format during the data processing pipeline.
\\
All 1200 images of the two datasets were then manually annotated by one person using the LabelMe tool with \acp{bbox} around the urine puddles and the label "puddle" \citep{Software_labelme}. In the process of annotation, the puddles are initially annotated in a first run and corrected in a second run by the precision of \acp{bbox} and the annotation of puddles that were not noticed in the first run. During the annotation of the datasets, only puddles of urine should be marked as the only object class. In some cases, urine and feces were also mixed. Particularly in the less sharp peripheral areas of the images away from the camera's focus, it was not possible to clearly differentiate between them. This can be seen in \autoref{fig:normal_vs_robust1}. As a result, feces left on the ground were also annotated as puddles. Another annotation challenge that occurred with the 773 images that had already been cut out is shown in \autoref{fig:normal_vs_robust3}. Here, it was difficult to differentiate between urine puddles that had already cooled down considerably and other similarly warm floor areas without the temporal progression from the video context. 

\noindent
The annotations are then transferred to the \ac{coco} format with the help of another program created in python and divided into the partial datasets for training, validation and testing. In addition to the top left corner and the width and height of the \acp{bbox}, the information about the area of the boxes is also included in the \ac{coco} formatted annotations. Very small puddles or splashes of urine are annotated as well at first. When loading the data during the training of the models, all objects with a height or width of 10 pixels or less are then filtered out. This is done in order to maintain a standardized procedure for borderline cases in object size. The filtered out objects are therefore not counted as puddles, but remain in the dataset. In addition, images that do not contain excretions and therefore have no annotated objects are specifically included in the datasets. The case that there are no excretions in the recording area of a camera occurs regularly and in a quantitatively relevant proportion, which means that these images should be included in a representative dataset. By learning that there does not have to be a excretion in every image, the AI models should achieve better generalization during training.
\\
The first dataset contains 1615 excretions, with no excretions occurring in 284 of the 1000 images. The second dataset contains 266 excretions, with no excretions in 68 of the 200 images. The storage space required, is 134 MB for the first and 22 MB for the second dataset. The second dataset differs significantly from the first dataset in some points. Due to the shorter distance of the camera to the ground, everything is slightly larger. This means that excretions of the same size and pigs occupy more pixels of the image in the second dataset than in the first dataset. As the camera in the second dataset recorded more directly from above, the excretions are covered more by the animals. Puddles of urine in the second dataset, which are created on the other floor, are significantly larger and take on a different shape and structure as they run off in one direction (\autoref{fig:normal_vs_robust2}). The size of the objects is also is defined as in the \ac{coco} dataset \citep{Software_COCO}. An object is considered small if it has a maximum size of $32*32=1024$ pixels. Above this, it is classified as medium-sized. Objects are considered large if they are larger than $96*96=9216$ pixels. While small objects dominate in the first dataset and no large objects are included, the second dataset contains few large objects and the ratio between small and medium-sized objects is more balanced. Finally, the images in the second dataset also contain elements that are not found in the first dataset. These include stripes on the ground heated by solar radiation and birds shown in \autoref{fig:normal_vs_robust4}, which can easily be mistaken for small excretions in the thermal images. In total \autoref{fig:normal_vs_robust} shows two annotated images from each of the two datasets and illustrates the differences between the datasets and the challenges during annotation.

\subsection{Models}
\label{subsec:models}
\noindent
Four different proven and state-of-the-art object detection models are tested and compared to each other. The models include the Faster \ac{r-cnn} as a two-stage detector \citep{Software_Faster_R-CNN}, the \ac{yolo} in the eighth version (v8) as a single-stage detector \citep{Software_YOLOv8} as well as \ac{detr} and \ac{dab-detr} as representatives for transformer-based models for object detection \citep{Software_DETR,Software_DAB-DETR}. 
%A schematic representation of the architectures described below can be found in \ref{appendix:A}.

\subsubsection{Faster R-CNN}
\noindent
The Faster \ac{r-cnn} is a model for object detection that uses \acp{cnn} and \ac{dl}. It can recognize multiple objects in an image and assign each to a class. Essentially, its structure can be divided into the sections Backbone, Region Proposal Network (RPN) and Region of Interest Pooling (RoI). The backbone is usually a \ac{cnn} and is used to extract features or characteristics from the images. The \ac{resnet} developed for image processing is used as a backbone, as it can deal well with the problem of vanishing gradients at deep \acp{ann} with the help of jump connections \citep{Software_ResNet,dynamische_NN}. The \ac{rpn} as an intermediate step makes the Faster \ac{r-cnn} a two-stage detector and therefore generally more accurate, but also slower than single-stage detectors.

\subsubsection{YOLOv8}
\noindent
The \ac{yolo}v8 developed by Ultralytics is a modern model for object detection \citep{Software_YOLOv8}. Based on \ac{yolo} and a continuous evolution over \ac{yolo}v3 and \ac{yolo}v5 to improve its performance and flexibility, \ac{yolo}v8 resulted in January 2023 \citep{Software_YOLO,Software_YOLOv3,Software_YOLOv5,Software_YOLOv8}  . The architecture of \ac{yolo}v8 consists of a backbone, neck and head, whereby the neck is responsible for linking various parts of the backbone to the head. A \ac{cnn} is also used as the backbone here. However, it is not a \ac{resnet}, but a CSPDarknet53. This combines the darknet53 with the \ac{csp} and was already used in \ac{yolo}v4 \citep{Software_YOLOv4, Software_CSPNet}. The classification of the objects and the regression of the \acp{bbox} takes place directly and without an intermediate step in the head, which defines the YOLOv8 as a single-stage detector. 

\subsubsection{DETR}
\noindent
A different approach is taken by \ac{detr}. Transformers were originally developed in 2017 for dealing with languages and linguistics \citep{Software_Transformer}. Their architecture consists of two multi-layer blocks, an encoder and a decoder. However, instead of words, the transformer in \ac{detr} receives the output of a \ac{cnn} applied to images as input \citep{Software_DETR}. Even though there are various options to choose from, a \ac{resnet} usually takes the place of the backbone. The transformer architecture enables \ac{detr} to recognize and consider global relationships between objects and to generate the output directly and in parallel. The model recognizes large objects better than a Faster \ac{r-cnn}, but is inferior to it in recognizing small objects \citep{Software_DETR}. 

\subsubsection{DAB-DETR}
\noindent
\ac{dab-detr} is based on \ac{detr}. In contrast, it uses dynamically updated anchor boxes instead of positional encodings. This improves the slow training convergence of \ac{detr} \citep{Software_DAB-DETR}.  \ac{dab-detr} requires fewer epochs for training due to the faster convergence with comparable or better results \citep{Software_DAB-DETR}. 

\subsection{Hyperparameters and model configuration}
\noindent
MMDetection and MMYOLO are used to create and configure the models. They are sub-projects of OpenMMLab, an open source project based on PyTorch and a developer community specializing in computer vision and machine learning \citep{Software_MMDetection-preprint,Software_MMDetection,Software_MMYOLO}. 
\\
All models are taken from the Model Zoo of MMDetection or MMYOLO, preconfigured for the \ac{coco} dataset \citep{Software_COCO}, and then were adapted to the excretion dataset and this work. However, only a few adjustments to the hyperparameters are tested as part of the work. Finally, the objectives of the work do not include a complete fine-tuning of the hyperparameters for the dataset. Rather they include a comparison of models and configurations, that have been proven on known and large datasets, on the specific task and dataset of this work. For all hyperparameters that are not varied in this work, such as the number of epochs in the training and the optimizers used, the assumption is made that they are also suitable and that an optimization would only have scaling and no ranking effects on the results of the models.  Furthermore, these hyperparameters are not synchronized between the models, as different values result in optimal performance for the different architectures of the models. It is precisely this best performance of the models on the dataset that is to be measured and then compared. Examples of such differences are the already mentioned number of training epochs for \ac{detr} and \ac{dab-detr} or the different size adjustments of the images depending on the model architecture.
\\
Hyperparameters like the used backbone, data preprocessing and augmentation, are also not varied, but are synchronized between the models for better comparability. For example, Faster R-CNN, DETR and DAB-DETR are all configured with a \ac{resnet}50 as backbone. A comparison of the most important properties in which the models differ is given in \autoref{tab:eigenschaften}. 

\begin{table}[h]
  \centering
  \caption[Comparison of important properties between the models]{Comparison of important different parameters and properties between the models. CSPDarknet53 abbreviated as CSPD53.}
  \label{tab:eigenschaften}
  \begin{tabular}{ccccc}
      \toprule
      Feature & Faster R-CNN & YOLOv8 & DETR & DAB-DETR \\
      \midrule
      Backbone & ResNet50 & CSPD53 & ResNet50 & ResNet50 \\ 
      Other parts & RPN, RoI & Neck,Head & En-/Decoder & En-/Decoder \\
      Parameters & 41,348M & 11,136M & 41,555M & 43,702M \\ 
      Epochs & 24 & 500 & 150 & 50 \\ 
      Opimizer & SDG & SDG & AdamW & AdamW \\ 
      \bottomrule
  \end{tabular}
\end{table}
\noindent
The same normalizations and the same filter options are used across all models when preprocessing the data. The methods used for augmentation, which are identical across all models, include horizontal and vertical mirroring with a probability of 20\% each. For a further 20\%, the images are rotated at a random angle across all models. An image is thus rotated or mirrored in one of the two directions by 60\% for all models. 
\\
Three hyperparameters considered especially related to the characteristics of the dataset are varied for all four models. In addition to the value from the model Zoo, a further value is tested in each case, which is assumed to fit the specific dataset of this work better than the value selected for the \ac{coco} dataset \citep{Software_COCO}. The batch size during training, the weighting of the class loss and the training status are varied. The corresponding values were determined empirically in preliminary tests with a subset of the first dataset. 
\\
The first excretion dataset created in this work with 1000 images is significantly smaller than the \ac{coco} dataset. This means significantly fewer iterations per epoch for the same batch size. Fewer iterations can easily lead to poorer generalization and more overfitting \citep{AI_large-batch_generalization}. Therefore, in addition to the adopted batch size of 16, a smaller batch size of 4 is tested. For \ac{yolo}v8, 16 is also used as the high value instead of the preconfigured batch size of 32, as the working memory of \acp{gpu} could otherwise reach its limits and the results are thus easier to compare. The learning rate is automatically adapted to the change in batch size using a linear scaling rule \citep{Software_linear-scaling}. 
\\
Since the excretions dataset only contains one object class, while the COCO dataset contains 80 classes, learning the correct class prediction is less complex. Therefore, in addition to the adopted weighting of the class loss, a weighting that is weaker by a factor of 100 is also tested. This means that the focus when training the models is much more on localizing the object positions and determining the size of the \acp{bbox}. The original weighting of the loss is 0,5 for \ac{yolo}v8 and 1,0 for the other three models. Accordingly, the reduced weightings are 0,005 and 0,01.
\\
In the \ac{coco} dataset, none of the 80 classes describe urine puddles, normal puddles or objects similar to puddles \citep{Software_COCO}, so the benefit of a model pre-trained entirely on the \ac{coco} dataset is unclear. To investigate this, the models are tested both pre-trained and untrained on the \ac{coco} dataset. For Faster-\ac{r-cnn}, \ac{detr} and \ac{dab-detr}, untrained means that only the \ac{resnet}50 backbone is pre-trained on the ImageNet dataset \citep{Software_ImageNet}, but not the other trainable components of the models, such as the \ac{rpn} or the encoder.  For \ac{yolo}v8, it means that it is completely untrained (including the backbone). Pre-trained, on the other hand, means that all trainable components of the models are loaded pre-trained on the COCO dataset \citep{Software_COCO}.
\\
For each model, each variant of a hyperparameter is tested in combination with each variant of the other two hyperparameters in the form of a grid search. With two variants per hyperparameter and three varied hyperparameters, this results in eight different \acp{hpc} per model, which are listed in \autoref{tab:kombinationen}. 

\begin{table}[h]
    \centering
    \caption{Composition of the hyperparmeter combinations used.}
    \label{tab:kombinationen}
    \begin{tabular}{cccc}
        \toprule
        Combination & training status & batch size & class loss weighting \\
        \midrule
        1 & Pre-trained (p) & 16 & original (L) \\
        2 & Pre-trained (p)& 16 & reduced (l)\\
        3 & Pre-trained (p)& 4 & original (L)\\
        4 & Pre-trained (p)& 4 & reduced (l)\\
        5 & Untrained (u)& 16 & original (L)\\
        6 & Untrained (u)& 16 & reduced (l)\\
        7 & Untrained (u)& 4 & original (L)\\
        8 & Untrained (u)& 4 & reduced (l)\\
        \bottomrule
    \end{tabular}
\end{table}
\noindent
The backbones of the models are designed for images with three channels, for instance from the RGB or HSV color space. However, grayscale images, such as those in this work, are not a combination of several channels, but have only one channel. To solve the problem of this discrepancy, the one channel is tripled when loading the datasets, resulting in images with three identical channels. This means that the backbones do not have to be changed. This also corresponds to the current standard handling of grayscale images by MMDetection and MMYOLO \citep{Software_MMDetection,Software_MMYOLO}.

\subsection{Training, validation and testing}
\label{subsec:train_val_test}
\noindent
The models are trained on an AI server. The \acp{gpu} used for this are NVIDIA A100 Tensor Core \acp{gpu} with 80 GB of RAM. One of these is divided into two instances with 40 GB of RAM each using \ac{mig}. This allows up to three models to be trained in parallel. 
\\
Each training runs at least until the metrics converge. This is checked manually by monitoring the progress of the metrics over the training epochs. The tool used for this is TensorBoard from TensorFlow \citep{SOftware_TensorFlow}. Due to strongly fluctuating metrics in the training process with DETR and only a few necessary training epochs with Faster-R-CNN, early stopping is not used as a method for convergence detection. Instead, the best epochs per cross-validation from the training, based on the validation, are saved and then tested with the test datasets.
\\
The models are trained with the first dataset consisting of 1000 images using a method derived from nested cross-validation. As with nested cross-validation, there is an inner and an outer loop. For the inner and outer loops we set k=5. In the outer loop, the test and training data are split. In the inner loop, the training data is then divided into the data for training and the data for validation. This results in a division into 640 images for training, 160 images for validation and 200 images for testing. In contrast to nested cross-validation, after the last run of the inner loop, no averaging of the five runs and a subsequent selection of the best \ac{hpc} takes place, which is then trained again with the 800 images for training and validation and tested with the 200 test images from the outer loop. Instead, all 5 results per \ac{hpc} are tested directly with the 200 images from the test dataset of the outer loop and the 200 images of the second test dataset from the freely ventilated barn for robustness. After completing the fifth and last outer loop, each \ac{hpc} was thus trained a total of 25 times with a different split of the data and tested with test datasets that were neither part of the training nor the validation. For each \ac{hpc}, 25 results are thus available for each of the two test datasets, which can then be averaged, selected and statistically evaluated. More detailed information on the deviations from the nested cross-validation as well as figures that visualize them are given in \ref{appendix:B}.

\subsection{Evaluation}
\label{subsec:evaluation}
\noindent
The \ac{ap} and the \ac{ar} are used as metrics for the assessment and evaluation of the model performance in this work. The \ac{ap} is raised for a threshold of \ac{iou} of 50\% (AP@50) and 75\% (AP@75) as well as averaged for the thresholds of \ac{iou} in the range from 50\% to 95\% (AP) at intervals of 5\%.  For the averaged thresholds, a distinction is also made between all objects, only medium-sized objects (APm) and only small objects (APs). As there are only a few large objects in the dataset, no metrics are collected specifically for them. The \ac{ar} is also averaged for thresholds over a range of \ac{iou} from 50\% to 95\% for all (AR), medium (ARm) and small (ARs) objects. These eight metrics are also part of the 12 metrics used for the evaluation of the \ac{coco} dataset \citep{Software_COCO}. A more detailed description of the 
basic evaluation metrics required to compute the \ac{ap} and \ac{ar} scores, their formulas and their interpretability in the context of this work is provided in \ref{appendix:C}. The F1 score is not calculated, as it cannot be assumed that precision and recall are equally important in the context of this study. The importance of precision and recall for the detection of excretions, the resulting consequences for practice and the generation of emissions are discussed in more detail in \autoref{sec:Discussion}.
\\
The results obtained from the test datasets are then used to calculate mean values for the individual metrics. The best \ac{hpc} per model and metric based on the mean value is then used for various statistical analyses. These analyses are carried out using Python version 3.10 and the numpy, pandas, scikit\_posthocs and scipy libraries.  First, the normal distribution of the data is checked using the Shapiro-Wilk test at a significance level of $\alpha=0.05$. Then, depending on the results of the Shapiro-Wilk test, a \ac{anova} or a Kruskal-Wallis test is performed for each metric to check for significant differences between the models in the metric. For both the \ac{anova} and the Kruskal-Wallis test, the significance level is $\alpha=0.05$. For metrics with significant differences between the models, these differences are then analyzed in pairs depending on the normal distribution of the data using T-tests or Dunn tests. For both tests, a Bonferroni correction is applied to counteract the alpha error accumulation of the multiple tests. The significance level for the pairwise tests is therefore $\alpha=0.05$ before Bonferroni correction and $\alpha=0.0083$ afterwards.

\section{Results}
\label{sec:Results}

\subsection{Training process}
\noindent
The time required for training differs for the models. Faster \ac{r-cnn} and \ac{dab-detr} needed about one minute for each epoch, \ac{dab-detr} about 40 seconds and \ac{yolo}v8 about 15 seconds. The mentioned duration per epoch refers to the \acp{hpc} with the smaller batch of four. Combinations with the larger batch were slightly faster for all models. However, since the models train over different numbers of epochs (see \autoref{tab:eigenschaften}), \ac{yolo}v8 required the most time for the whole training, followed by \ac{detr} and \ac{dab-detr}. The Faster \ac{r-cnn} was the fastest model in the entire training. The trained models then took about 15 seconds for Faster \ac{r-cnn} and \ac{dab-detr}, about 12 seconds for \ac{detr} and about 5 seconds for \ac{yolo}v8 for a test on the test datasets with 200 images each. The training process monitored with Tensorboard shows converging metrics and a converging loss for Faster \ac{r-cnn}. However, the \acp{hpc} with the reduced class loss weighting show an increase in loss before the loss decreases and converges. The metrics and the loss also converge for \ac{yolo}v8. The metrics begin to converge between epoch 200 and 300. The loss decreases and converges here for all \acp{hpc}. For both models, the convergence of the loss from the combinations with the batch of 16 is just beginning, while for the smaller batch the convergence is much more significant. The training curve of \ac{detr} shows an early convergence of the loss. The untrained combinations remain at a high loss value, which is about three times the loss of the pre-trained models. The associated metrics of these models show no improvement over the training process. For \ac{ar}, all other combinations show a rapid increase and convergence. The combinations of the smaller batch with the reduced loss weighting, on the other hand, show only a slight increase in the \ac{ap} metrics at the beginning and then convergence at a very low level. The \ac{dab-detr}, whose architecture is a further development of the \ac{detr} architecture, also shows some parallels in the training process, but with deviations. The loss of the untrained combinations is also around three times higher than for the untrained models. However, the two untrained combinations with the original loss decrease without any signs of convergence. Convergence of the loss can be seen in the other six variants. Looking at the metrics during the training of \ac{dab-detr}, the untrained variants of the model with the smaller loss weighting do not improve, while with the original weighting a flat increase in the metrics begins in the second half of the training. The variants pre-trained with the \ac{coco} dataset have high values in the \ac{ar} metrics from the beginning, improve slightly and then converge. In the \ac{ap} metrics, on the other hand, they start very low, then increase to varying degrees and for a long time until they also converge.

\begin{table}[h]
    \centering
    \caption[Average best training epoch]{Average and standard deviation of the best training epochs according to \ac{ap} per \ac{hpc} and model. The \acp{hpc} are given in the form batch size\_loss weighting\_training status, where \textbf{L} stands for the original and \textbf{l} for the 100 times reduced weighting, as well as \textbf{p} for pretrained and \textbf{u} for untrained.}
    \label{tab:epochen}
    \begin{adjustbox}{max width=\textwidth}
    \begin{tabular}{lcccc}
      \toprule
      \textbf{HPC} & \textbf{Faster R-CNN} & \textbf{YOLOv8} & \textbf{DETR} & \textbf{DAB-DETR} \\
      \midrule
      \textbf{4\_L\_p} & 20,16 $\pm$ 2.63 & 190,52 $\pm$ 94,52 & 134,72 $\pm$ 12,62 & 47,24 $\pm$ 2,08 \\
      \textbf{4\_L\_u} & 22,20 $\pm$ 1,62 & 433,52 $\pm$ 52,51 & 23,64 $\pm$ 34,07 & 46,00 $\pm$ 3,24 \\
      \textbf{4\_l\_p} & 21,36 $\pm$ 1,79 & 317,92 $\pm$ 95,81 & 99,88 $\pm$ 32,56 & 47,28 $\pm$ 3,97 \\
      \textbf{4\_l\_u} & 21,68 $\pm$ 2,20 & 464,00 $\pm$ 36,71 & 53,76 $\pm$ 48,24 & 1,00 $\pm$ 0,00 \\
      \textbf{16\_L\_p} & 21,04 $\pm$ 2,41 & 184,88 $\pm$ 52,44 & 136,44 $\pm$ 11,87 & 48,36 $\pm$ 1,55 \\
      \textbf{16\_L\_u} & 22,68 $\pm$ 1,32 & 324,52 $\pm$ 72,05 & 44,04 $\pm$ 52,75 & 44,24 $\pm$ 9,99 \\
      \textbf{16\_l\_p} & 21,88 $\pm$ 2,12 & 297,88 $\pm$ 90,05 & 135,92 $\pm$ 9,94 & 48,64 $\pm$ 1,26 \\
      \textbf{16\_l\_u} & 22,68 $\pm$ 1,32 & 435,24 $\pm$ 53,35 & 33,40 $\pm$ 40,71 & 2,16 $\pm$ 5,68 \\
      \bottomrule
    \end{tabular}
    \end{adjustbox}
\end{table}
\noindent
The training epoch with the best \ac{ap} in the validation, which was subsequently tested with the two test datasets, varies between the \acp{hpc}. In some cases, large deviations also occur within a \ac{hpc}. An overview of this is provided by \autoref{tab:epochen}, which contains the average best training epoch for each model and each \ac{hpc} including the associated standard deviation. Faster \ac{r-cnn} and \ac{yolo}v8 therefore reached the epoch with the best \ac{ap} later in training when untrained than the corresponding pre-trained variants. For \ac{yolo}v8, this also applies to the variants with reduced compared to original loss weighting. For \ac{detr} and \ac{dab-detr}, the pattern between untrained and pre-trained variants can be described inversely to Faster \ac{r-cnn} and \ac{yolo}v8. Similarly, in the third column of the table (for \ac{detr}), the reverse pattern of reduced and original loss weighting compared to \ac{yolo}v8 can be described.

\subsection{Hyperparameters and models}

\begin{table}[h]
    \centering
    \caption[Results of Faster R-CNN]{Results of various \ac{ap} and \ac{ar} metrics of Faster \ac{r-cnn} on the first test dataset. The \acp{hpc} are given in the form batch-size\_loss-weighting\_training-status, where \textbf{L} stands for the original and \textbf{l} for the 100-times reduced weighting, as well as \textbf{p} for pretrained and \textbf{u} for untrained.}
    \label{tab:ergebnisse-faster_r-cnn}
    \begin{adjustbox}{max width=\textwidth}
    \begin{tabular}{lcccccccc}
        \toprule
        \textbf{Faster R-CNN} & \textbf{4\_L\_p} & \textbf{4\_L\_u} & \textbf{4\_l\_p} & \textbf{4\_l\_u} & \textbf{16\_L\_p} & \textbf{16\_L\_u} & \textbf{16\_l\_p} & \textbf{16\_l\_u} \\
        \midrule
        %AP & \textbf{0,581} & 0,548 & 0,517 & 0,419 & 0,579 & 0,536 & 0,498 & 0,357 \\
        %AP@50 & \textbf{0,919} & 0,916 & 0,825 & 0,733 & 0,918 & 0,913 & 0,803 & 0,657 \\
        %AP@75 & \textbf{0,678} & 0,606 & 0,596 & 0,448 & 0,672 & 0,582 & 0,574 & 0,348 \\
        %APs & \textbf{0,561} & 0,533 & 0,501 & 0,416 & 0,560 & 0,522 & 0,487 & 0,363 \\
        %APm & \textbf{0,694} & 0,631 & 0,647 & 0,539 & 0,684 & 0,615 & 0,629 & 0,461 \\
        %AR & 0,651 & 0,622 & \textbf{0,659} & 0,595 & 0,650 & 0,611 & 0,656 & 0,559 \\
        %ARs & 0,634 & 0,610 & \textbf{0,645} & 0,587 & 0,634 & 0,601 & 0,643 & 0,555 \\
        %ARm & \textbf{0,747} & 0,690 & 0,738 & 0,645 & 0,738 & 0,672 & 0,730 & 0,586 \\
        AP in \% & \textbf{58,1}$\pm$2,3 & 54,8$\pm$1,7 & 51,7$\pm$2,6 & 41,9$\pm$1,9 & 57,9$\pm$1,9 & 53,6$\pm$1,8 & 49,8$\pm$2,4 & 35,7$\pm$2,9 \\
        AP@50 in \% & \textbf{91,9}$\pm$1,6 & 91,6$\pm$1,1 & 82,5$\pm$2,6 & 73,3$\pm$2,5 & 91,8$\pm$1,5 & 91,3$\pm$1,4 & 80,3$\pm$2,9 & 65,7$\pm$4,6 \\
        AP@75 in \% & \textbf{67,8}$\pm$4,2 & 60,6$\pm$4,3 & 59,6$\pm$4,1 & 44,8$\pm$3,2 & 67,2$\pm$3,1 & 58,2$\pm$3,0 & 57,4$\pm$4,0 & 34,8$\pm$3,7 \\
        APs in \% & \textbf{56,1}$\pm$2,0 & 53,3$\pm$1,9 & 50,1$\pm$2,4 & 41,6$\pm$1,7 & 56,0$\pm$1,5 & 52,2$\pm$1,8 & 48,7$\pm$2,5 & 36,3$\pm$2,7 \\
        APm in \% & \textbf{69,4}$\pm$4,0 & 63,1$\pm$2,5 & 64,7$\pm$4,2 & 53,9$\pm$3,7 & 68,4$\pm$3,6 & 61,5$\pm$2,9 & 62,9$\pm$4,5 & 46,1$\pm$5,4 \\
        AR in \% & \textbf{65,1}$\pm$1,9 & 62,2$\pm$1,5 & \textbf{65,9}$\pm$1,9 & 59,5$\pm$1,6 & 65,0$\pm$1,6 & 61,1$\pm$1,5 & 65,6$\pm$1,9 & 55,9$\pm$1,7 \\
        ARs in \% & \textbf{63,4}$\pm$1,8 & 61,0$\pm$1,5 & \textbf{64,5}$\pm$1,6 & 58,7$\pm$1,4 & 63,4$\pm$1,4 & 60,1$\pm$1,4 & 64,3$\pm$1,6 & 55,5$\pm$1,8 \\
        ARm in \% & \textbf{74,7}$\pm$3,1 & 69,0$\pm$2,7 & 73,8$\pm$4,2 & 64,5$\pm$4,1 & 73,8$\pm$3,1 & 67,2$\pm$3,1 & 73,0$\pm$4,1 & 58,6$\pm$4,4 \\
        \bottomrule
    \end{tabular}
    \end{adjustbox}
\end{table}
\noindent
\autoref{tab:ergebnisse-faster_r-cnn} shows the mean values per metric and \ac{hpc} of the Faster \ac{r-cnn} on the first test dataset. As in the tables for the other models, the value of the best \ac{hpc} for each metric is highlighted in bold. For all \ac{ap} metrics as well as for the \ac{ar} over the medium sized objects, the combination of the small batch size of 4 with the original loss weighting and the completely pre-trained model on the \ac{coco} dataset was the best. Closely followed by the same combination with the original batch size of 16. For the \ac{ar} over all objects and the small objects, however, the combination of the smaller batch size with the 100-times smaller class loss and the completely pre-trained model was the best.  The values of the combinations with the larger batch are all slightly below or at the same level as the values of their variants with the smaller batch. In \autoref{tab:ergebnisse-yolo_v8} you can see the mean values of the test results of the \ac{yolo}v8 with the first test dataset. The results can be described identically to those of the Faster \ac{r-cnn}. Only in the \ac{ap}@50 the combination with a batch of 16, the original weighting of the class loss and the pre-trained model ist equal to the same combination with a batch of 4. For the \ac{yolo}v8, the gap between pre-trained and untrained combinations is smaller than for the other three models. Especially in the metrics of \ac{ar}, all eight combinations are at a very similar level. All four untrained combinations with a small batch size are inferior to those with a large batch size.

\begin{table}[h]
    \centering
    \caption[Results of YOLOv8]{Results of various \ac{ap} and \ac{ar} metrics of \ac{yolo}v8 on the first test dataset. The \acp{hpc} are given in the form batch size\_loss weighting\_training status, where \textbf{L} stands for the original and \textbf{l} for the 100-times reduced weighting, as well as \textbf{p} for pretrained and \textbf{u} for untrained.}
    \label{tab:ergebnisse-yolo_v8}
    \begin{adjustbox}{max width=\textwidth}
    \begin{tabular}{lcccccccc}
        \toprule
        \textbf{YOLOv8} & \textbf{4\_L\_p} & \textbf{4\_L\_u} & \textbf{4\_l\_p} & \textbf{4\_l\_u} & \textbf{16\_L\_p} & \textbf{16\_L\_u} & \textbf{16\_l\_p} & \textbf{16\_l\_u} \\
        \midrule
        %AP & \textbf{0,593} & 0,561 & 0,563 & 0,499 & 0,584 & 0,573 & 0,562 & 0,516 \\
        %AP@50 & \textbf{0,908} & 0,900 & 0,871 & 0,793 & \textbf{0,908} & 0,907 & 0,874 & 0,805 \\
        %AP@75 & \textbf{0,686} & 0,639 & 0,647 & 0,573 & 0,674 & 0,654 & 0,652 & 0,602 \\
        %APs & \textbf{0,560} & 0,531 & 0,529 & 0,469 & 0,554 & 0,544 & 0,531 & 0,485 \\
        %APm & \textbf{0,744} & 0,702 & 0,715 & 0,650 & 0,725 & 0,711 & 0,711 & 0,670 \\
        %AR & 0,670 & 0,651 & \textbf{0,673} & 0,656 & 0,662 & 0,657 & 0,670 & 0,664 \\
        %ARs & 0,649 & 0,632 & \textbf{0,654} & 0,639 & 0,641 & 0,638 & 0,650 & 0,646 \\
        %ARm & \textbf{0,787} & 0,755 & 0,778 & 0,752 & 0,775 & 0,763 & 0,781 & 0,766 \\
         AP in \% & \textbf{59,3$\pm$2,2} & 56,1$\pm$2,2 & 56,3$\pm$2,9 & 49,9$\pm$2,5 & 58,4$\pm$2,1 & 57,3$\pm$2,0 & 56,2$\pm$2,7 & 51,6$\pm$2,9 \\
        AP@50 in \% & \textbf{90,8$\pm$1,1} & 90,0$\pm$1,7 & 87,1$\pm$2,3 & 79,3$\pm$2,6 & \textbf{90,8$\pm$1,3} & 90,7$\pm$1,8 & 87,4$\pm$2,3 & 80,5$\pm$3,0 \\
        AP@75 in \% & \textbf{68,6$\pm$4,3} & 63,9$\pm$4,1 & 64,7$\pm$4,0 & 57,3$\pm$4,3 & 67,4$\pm$3,9 & 65,4$\pm$3,7 & 65,2$\pm$4,4 & 60,2$\pm$4,4 \\
        APs in \% & \textbf{56,0$\pm$2,3} & 53,1$\pm$2,4 & 52,9$\pm$2,8 & 46,9$\pm$2,4 & 55,4$\pm$2,1 & 54,4$\pm$2,2 & 53,1$\pm$2,6 & 48,5$\pm$3,0 \\
        APm in \% & \textbf{74,4$\pm$2,5} & 70,2$\pm$2,2 & 71,5$\pm$3,7 & 65,0$\pm$4,7 & 72,5$\pm$2,8 & 71,1$\pm$2,2 & 71,1$\pm$3,9 & 67,0$\pm$3,8 \\
        AR in \% & 67,0$\pm$2,0 & 65,1$\pm$1,9 & \textbf{67,3$\pm$2,1} & 65,6$\pm$2,1 & 66,2$\pm$2,1 & 65,7$\pm$1,9 & 67,0$\pm$1,8 & 66,4$\pm$1,8 \\
        ARs in \% & 64,9$\pm$2,0 & 63,2$\pm$2,0 & \textbf{65,4$\pm$2,1} & 63,9$\pm$2,1 & 64,1$\pm$2,1 & 63,8$\pm$2,0 & 65,0$\pm$1,8 & 64,6$\pm$1,9 \\
        ARm in \% & \textbf{78,7$\pm$2,4} & 75,5$\pm$1,9 & 77,8$\pm$2,4 & 75,2$\pm$3,0 & 77,5$\pm$2,4 & 76,3$\pm$2,3 & 78,1$\pm$2,2 & 76,6$\pm$2,0 \\
        \bottomrule
    \end{tabular}
    \end{adjustbox}
\end{table}
 
\begin{table}[h]
    \centering
    \caption[Results of DETR]{Results of various \ac{ap} and \ac{ar} metrics of \ac{detr} on the first test dataset. The \acp{hpc} are given in the form batch size\_loss weighting\_training status, where \textbf{L} stands for the original and \textbf{l} for the 100-times reduced weighting, as well as \textbf{p} for pretrained and \textbf{u} for untrained.}
    \label{tab:ergebnisse-detr}
    \begin{adjustbox}{max width=\textwidth}
    \begin{tabular}{lcccccccc}
        \toprule
        \textbf{DETR} & \textbf{4\_L\_p} & \textbf{4\_L\_u} & \textbf{4\_l\_p} & \textbf{4\_l\_u} & \textbf{16\_L\_p} & \textbf{16\_L\_u} & \textbf{16\_l\_p} & \textbf{16\_l\_u} \\
        \midrule
        %AP & 0,471 & 0,000 & 0,053 & 0,000 & \textbf{0,479} & 0,000 & 0,396 & 0,000 \\
        %AP@50 & 0,871 & 0,000 & 0,094 & 0,000 & \textbf{0,875} & 0,000 & 0,698 & 0,000 \\
        %AP@75 & 0,450 & 0,000 & 0,054 & 0,000 & \textbf{0,466} & 0,000 & 0,414 & 0,000 \\
        %APs & 0,433 & 0,000 & 0,049 & 0,000 & \textbf{0,440} & 0,000 & 0,343 & 0,000 \\
        %APm & \textbf{0,647} & 0,000 & 0,097 & 0,000 & 0,646 & 0,000 & 0,633 & 0,000 \\
        %AR & 0,608 & 0,001 & \textbf{0,662} & 0,001 & 0,610 & 0,000 & 0,659 & 0,000 \\
        %ARs & 0,588 & 0,000 & 0,638 & 0,001 & 0,591 & 0,000 & \textbf{0,641} & 0,000 \\
        %Rm & 0,719 & 0,002 & \textbf{0,798} & 0,003 & 0,714 & 0,001 & 0,757 & 0,001 \\
        AP in \% & 47,1$\pm$2,0 & 0,0$\pm$0,0 & 5,3$\pm$1,1 & 0,0$\pm$0,1 & \textbf{47,9$\pm$2,1} & 0,0$\pm$0,0 & 39,6$\pm$3,0 & 0,0$\pm$0,0 \\
        AP@50 in \% & 87,1$\pm$2,0 & 0,0$\pm$0,0 & 9,4$\pm$1,6 & 0,0$\pm$0,2 & \textbf{87,5$\pm$1,7} & 0,0$\pm$0,0 & 69,8$\pm$4,5 & 0,0$\pm$0,0 \\
        AP@75 in \% & 45,0$\pm$3,9 & 0,0$\pm$0,0 & 5,4$\pm$1,5 & 0,0$\pm$0,0 & \textbf{46,6$\pm$4,2} & 0,0$\pm$0,0 & 41,4$\pm$4,7 & 0,0$\pm$0,0 \\
        APs in \% & 43,3$\pm$1,8 & 0,0$\pm$0,1 & 4,9$\pm$1,0 & 0,0$\pm$0,1 & \textbf{44,0$\pm$1,9} & 0,0$\pm$0,0 & 34,3$\pm$2,8 & 0,0$\pm$0,1 \\
        APm in \% & \textbf{64,7$\pm$3,3} & 0,0$\pm$0,0 & 9,7$\pm$2,6 & 0,0$\pm$0,2 & 64,6$\pm$3,4 & 0,0$\pm$0,1 & 63,3$\pm$4,0 & 0,0$\pm$0,1 \\
        AR in \% & 60,8$\pm$1,6 & 0,1$\pm$0,1 & \textbf{66,2$\pm$3,2} & 0,1$\pm$0,2 & 61,0$\pm$1,6 & 0,0$\pm$0,1 & 65,9$\pm$1,8 & 0,0$\pm$0,1 \\
        ARs in \% & 58,8$\pm$1,5 & 0,0$\pm$0,1 & 63,8$\pm$3,5 & 0,1$\pm$0,1 & 59,1$\pm$1,4 & 0,0$\pm$0,1 & \textbf{64,1$\pm$1,7} & 0,0$\pm$0,1 \\
        ARm in \% & 71,9$\pm$2,6 & 0,2$\pm$0,4 & \textbf{79,8$\pm$3,4} & 0,3$\pm$0,5 & 71,4$\pm$2,8 & 0,1$\pm$0,3 & 75,7$\pm$2,8 & 0,1$\pm$0,1 \\
        \bottomrule
    \end{tabular}
    \end{adjustbox}
\end{table}
\noindent
The averaged results of the \ac{detr} on the first dataset are shown in \autoref{tab:ergebnisse-detr}. First of all, it is noticeable that all untrained combinations are equal or close to zero across all metrics. The pre-trained combination with the original loss weighting and the batch of 16 resulted in the best values for all \ac{ap} metrics with the exception of \ac{ap}m. For \ac{ap}m, the corresponding combination with the small batch is slightly better on the average. Looking at the metrics of \ac{ar}, the pre-trained combinations with the smaller weighting of the class loss show the better performance. A larger batch is better for the small objects and the smaller batch is better for the medium-sized objects and the \ac{ar} across all object sizes. The determined metrics of the \ac{dab-detr} are listed in \autoref{tab:ergebnisse-dab-detr}. The \ac{hpc} in the first column, consisting of the smaller batch, the original loss weight and the pre-trained model, achieved the highest mean values in all \ac{ap} metrics. Similar to the other three models, the pre-trained \acp{hpc} with the reduced loss weight again showed the highest values for the \ac{ar}.  In the detection of small objects, the combination with the batch of 16 performs better, while the combination with the smaller batch is ahead in the detection of medium-sized objects. Across all objects both variants are equal. Parallel to the results of the \ac{detr}, considerably smaller values close to zero occur across all metrics in the columns of the untrained combinations.

\begin{table}[h]
    \centering
    \caption[Results of DAB-DETR]{Results of various \ac{ap} and \ac{ar} metrics of \ac{dab-detr} on the first test dataset. The \acp{hpc} are given in the form batch-size\_loss-weighting\_training-status, where \textbf{L} stands for the original and \textbf{l} for the 100-times reduced weighting, as well as \textbf{p} for pre-trained and \textbf{u} for untrained.}
    \label{tab:ergebnisse-dab-detr}
    \begin{adjustbox}{max width=\textwidth}
    \begin{tabular}{lcccccccc}
        \toprule
        \textbf{DAB-DETR} & \textbf{4\_L\_p} & \textbf{4\_L\_u} & \textbf{4\_l\_p} & \textbf{4\_l\_u} & \textbf{16\_L\_p} & \textbf{16\_L\_u} & \textbf{16\_l\_p} & \textbf{16\_l\_u} \\
        \midrule
        %AP & \textbf{0,510} & 0,014 & 0,403 & 0,000 & 0,506 & 0,008 & 0,465 & 0,000 \\
        %AP@50 & \textbf{0,898} & 0,064 & 0,709 & 0,000 & 0,895 & 0,037 & 0,809 & 0,000 \\
        %AP@75 & \textbf{0,525} & 0,002 & 0,416 & 0,000 & 0,509 & 0,001 & 0,481 & 0,000 \\
        %APs & \textbf{0,478} & 0,007 & 0,381 & 0,000 & 0,472 & 0,006 & 0,435 & 0,000 \\
        %APm & \textbf{0,666} & 0,073 & 0,586 & 0,000 & 0,663 & 0,031 & 0,647 & 0,000 \\
        %AR & 0,636 & 0,111 & \textbf{0,673} & 0,004 & 0,642 & 0,092 & \textbf{0,673} & 0,003 \\
        %ARs & 0,615 & 0,084 & 0,652 & 0,003 & 0,623 & 0,074 & \textbf{0,653} & 0,002 \\
        %ARm & 0,750 & 0,262 & \textbf{0,786} & 0,010 & 0,747 & 0,193 & 0,780 & 0,005 \\
        AP in \% & \textbf{51,0$\pm$1,4} & 1,4$\pm$0,7 & 40,3$\pm$5,0 & 0,0$\pm$0,0 & 50,6$\pm$1,9 & 0,8$\pm$0,5 & 46,5$\pm$2,5 & 0,0$\pm$0,0 \\
        AP@50 in \% & \textbf{89,8$\pm$1,8} & 6,4$\pm$2,8 & 70,9$\pm$9,1 & 0,0$\pm$0,0 & 89,5$\pm$1,8 & 3,7$\pm$2,2 & 80,9$\pm$2,8 & 0,0$\pm$0,0 \\
        AP@75 in \% & \textbf{52,5$\pm$2,6} & 0,2$\pm$0,2 & 41,6$\pm$6,1 & 0,0$\pm$0,0 & 50,9$\pm$3,8 & 0,1$\pm$0,1 & 48,1$\pm$4,2 & 0,0$\pm$0,0 \\
        APs in \% & \textbf{47,8$\pm$1,3} & 0,7$\pm$0,3 & 38,1$\pm$4,7 & 0,0$\pm$0,0 & 47,2$\pm$1,7 & 0,6$\pm$0,4 & 43,5$\pm$2,3 & 0,0$\pm$0,0 \\
        APm in \% & \textbf{66,6$\pm$2,6} & 7,3$\pm$2,9 & 58,6$\pm$7,2 & 0,0$\pm$0,0 & 66,3$\pm$3,8 & 3,1$\pm$2,6 & 64,7$\pm$4,3 & 0,0$\pm$0,0 \\
        AR in \% & 63,6$\pm$1,5 & 11,1$\pm$2,9 & \textbf{67,3$\pm$2,5} & 0,4$\pm$0,3 & 64,2$\pm$2,1 & 9,2$\pm$4,7 & \textbf{67,3$\pm$2,0} & 0,3$\pm$0,3 \\
        ARs in \% & 61,5$\pm$1,4 & 8,4$\pm$2,7 & 65,2$\pm$2,7 & 0,3$\pm$0,2 & 62,3$\pm$1,9 & 7,4$\pm$4,2 & \textbf{65,3$\pm$1,9} & 0,2$\pm$0,2 \\
        ARm in \% & 75,0$\pm$2,7 & 26,2$\pm$5,5 & \textbf{78,6$\pm$2,6} & 1,0$\pm$0,9 & 74,7$\pm$4,0 & 19,3$\pm$8,9 & 78,0$\pm$3,3 & 0,5$\pm$1,2 \\
        \bottomrule
    \end{tabular}
    \end{adjustbox}
\end{table}
\noindent
The next four tables contain the mean values of the results from the same models and metrics as the previous four tables, but on the second test dataset for robustness.  

\begin{table}[h]
    \centering
    \caption[Results for the robustness of the Faster R-CNN]{Results of various \ac{ap} and \ac{ar} metrics of the Faster \ac{r-cnn} on the second test dataset for robustness. The \acp{hpc} are given in the form batch-size\_loss-weighting\_training-status, where \textbf{L} stands for the original and \textbf{l} for the 100-times reduced weighting, as well as \textbf{p} for pre-trained and \textbf{u} for untrained.}
    \label{tab:ergebnisse-faster_r-cnn-2}
    \begin{adjustbox}{max width=\textwidth}
    \begin{tabular}{lcccccccc}
        \toprule
        \textbf{Faster R-CNN} & \textbf{4\_L\_p} & \textbf{4\_L\_u} & \textbf{4\_l\_p} & \textbf{4\_l\_u} & \textbf{16\_L\_p} & \textbf{16\_L\_u} & \textbf{16\_l\_p} & \textbf{16\_l\_u} \\
        \midrule
        %AP & 0,368 & 0,307 & 0,317 & 0,260 & \textbf{0,373} & 0,284 & 0,308 & 0,236 \\
        %AP@50 & \textbf{0,802} & 0,774 & 0,713 & 0,648 & 0,801 & 0,752 & 0,693 & 0,592 \\
        %AP@75 & 0,224 & 0,119 & 0,184 & 0,094 & \textbf{0,241} & 0,097 & 0,178 & 0,085 \\
        %APs & 0,378 & 0,318 & 0,352 & 0,287 & \textbf{0,386} & 0,293 & 0,346 & 0,268 \\
        %APm & \textbf{0,377} & 0,334 & 0,289 & 0,246 & 0,373 & 0,320 & 0,267 & 0,204 \\
        %AR & 0,449 & 0,389 & 0,491 & 0,425 & 0,463 & 0,375 & \textbf{0,492} & 0,425 \\
        %ARs & 0,457 & 0,395 & 0,519 & 0,454 & 0,475 & 0,379 & \textbf{0,522} & 0,461 \\
        %ARm & 0,469 & 0,436 & \textbf{0,473} & 0,394 & 0,468 & 0,434 & 0,464 & 0,361 \\
        AP in \% & 36,8$\pm$2,1 & 30,7$\pm$2,4 & 31,7$\pm$2,0 & 26,0$\pm$2,0 & \textbf{37,3$\pm$2,6} & 28,4$\pm$2,7 & 30,8$\pm$1,7 & 23,6$\pm$2,0 \\
        AP@50 in \% & \textbf{80,2$\pm$2,1} & 77,4$\pm$2,7 & 71,3$\pm$1,8 & 64,8$\pm$3,1 & 80,1$\pm$1,9 & 75,2$\pm$3,1 & 69,3$\pm$2,2 & 59,2$\pm$4,5 \\
        AP@75 in \% & 22,4$\pm$6,1 & 11,9$\pm$3,7 & 18,4$\pm$4,6 & 9,4$\pm$2,7 & \textbf{24,1$\pm$6,5} & 9,7$\pm$3,9 & 17,8$\pm$3,1 & 8,5$\pm$2,3 \\
        APs in \% & 37,8$\pm$2,3 & 31,8$\pm$2,7 & 35,2$\pm$2,4 & 28,7$\pm$2,2 & \textbf{38,6$\pm$3,0} & 29,3$\pm$3,1 & 34,6$\pm$1,9 & 26,8$\pm$2,3 \\
        APm in \% & \textbf{37,7$\pm$3,4} & 33,4$\pm$3,1 & 28,9$\pm$2,4 & 24,6$\pm$3,2 & 37,3$\pm$3,5 & 32,0$\pm$2,9 & 26,7$\pm$2,4 & 20,4$\pm$2,4 \\
        AR in \% & 44,9$\pm$2,4 & 38,9$\pm$2,3 & 49,1$\pm$2,0 & 42,5$\pm$2,5 & 46,3$\pm$2,6 & 37,5$\pm$3,1 & \textbf{49,2$\pm$1,4} & 42,5$\pm$1,8 \\
        ARs in \% & 45,7$\pm$2,6 & 39,5$\pm$2,8 & 51,9$\pm$2,6 & 45,4$\pm$2,9 & 47,5$\pm$3,0 & 37,9$\pm$3,7 & \textbf{52,2$\pm$1,8} & 46,1$\pm$2,3 \\
        ARm in \% & 46,9$\pm$3,3 & 43,6$\pm$3,2 & \textbf{47,3$\pm$2,2} & 39,4$\pm$3,6 & 46,8$\pm$3,3 & 43,4$\pm$2,9 & 46,4$\pm$3,0 & 36,1$\pm$3,4 \\
        \bottomrule
    \end{tabular}
    \end{adjustbox}
\end{table}
\noindent
\autoref{tab:ergebnisse-faster_r-cnn-2} shows the results of the Faster \ac{r-cnn} on the second dataset. In the metrics of \ac{ap}, the pre-trained variants with the original weighting of the class loss are the best. For the medium-sized objects and with a \ac{iou} threshold of 50\%, the combination with the smaller batch is slightly ahead of the otherwise leading combination with the batch of 16 images per iteration.  In the three bottom rows of the table, in which the values of the \ac{ar} are shown, the pretrained combinations with the reduced weighting of the loss have the highest performance.  Below this, the variant with the smaller batch leads for the medium-sized objects and the variant with the larger batch leads for the small objects and across all objects. The values of \ac{yolo}v8 on the second test dataset given in \autoref{tab:ergebnisse-yolo_v8-2} are almost exclusively highest in the first column from the left. This corresponds to the combination of the three varied hyperparameters with the smaller batch size of 4 images per iteration, the original weighting of the class loss and the use of a model pre-trained on the \ac{coco} dataset. The values of the other seven possible combinations are in some cases clearly below the values of this combination. Only the \ac{ar}s is minimally higher as the only metric in the third column from the left, i.e. with the smaller loss weighting. Compared to the values of the model with the first dataset in \autoref{tab:ergebnisse-yolo_v8}, the average difference to the values with the second dataset is the highest of the four models. Especially in \ac{ap}@75, a large difference to the first dataset can be seen for some of the \acp{hpc} with values below 10 \%. 

\begin{table}[h!]
    \centering
    \caption[Results for the robustness of the YOLOv8]{Results of various \ac{ap} and \ac{ar} metrics of the \ac{yolo}v8 on the second test dataset for robustness. The \acp{hpc} are given in the form batch size\_loss-weighting\_training status, where \textbf{L} stands for the original and \textbf{l} for the 100-times reduced weighting, as well as \textbf{p} for pretrained and \textbf{u} for untrained.}
    \label{tab:ergebnisse-yolo_v8-2}
    \begin{adjustbox}{max width=\textwidth}
    \begin{tabular}{lcccccccc}
        \toprule
        \textbf{YOLOv8} & \textbf{4\_L\_p} & \textbf{4\_L\_u} & \textbf{4\_l\_p} & \textbf{4\_l\_u} & \textbf{16\_L\_p} & \textbf{16\_L\_u} & \textbf{16\_l\_p} & \textbf{16\_l\_u} \\
        \midrule
        %AP & \textbf{0,298} & 0,129 & 0,228 & 0,097 & 0,262 & 0,158 & 0,212 & 0,164 \\
        %AP@50 & \textbf{0,778} & 0,527 & 0,643 & 0,386 & 0,724 & 0,594 & 0,641 & 0,536 \\
        %AP@75 & \textbf{0,140} & 0,010 & 0,084 & 0,012 & 0,108 & 0,017 & 0,064 & 0,042 \\
        %APs & \textbf{0,310} & 0,132 & 0,250 & 0,093 & 0,273 & 0,162 & 0,222 & 0,164 \\
        %APm & \textbf{0,317} & 0,163 & 0,225 & 0,140 & 0,275 & 0,186 & 0,224 & 0,203 \\
        %AR & \textbf{0,413} & 0,247 & 0,401 & 0,255 & 0,372 & 0,275 & 0,370 & 0,340 \\
        %ARs & 0,402 & 0,224 & \textbf{0,403} & 0,241 & 0,360 & 0,254 & 0,363 & 0,331 \\
        %ARm & \textbf{0,486} & 0,376 & 0,439 & 0,361 & 0,452 & 0,397 & 0,438 & 0,436 \\
        AP in \% & \textbf{29,8$\pm$2,6} & 12,9$\pm$2,2 & 22,8$\pm$2,5 & 9,7$\pm$4,4 & 26,2$\pm$5,0 & 15,8$\pm$2,7 & 21,2$\pm$2,9 & 16,4$\pm$2,3 \\
        AP@50 in \% & \textbf{77,8$\pm$2,8} & 52,7$\pm$5,7 & 64,3$\pm$3,6 & 38,6$\pm$12,1 & 72,4$\pm$8,0 & 59,4$\pm$5,4 & 64,1$\pm$4,1 & 53,6$\pm$5,8 \\
        AP@75 in \% & \textbf{14,0$\pm$3,5} & 1,0$\pm$0,8 & 8,4$\pm$3,3 & 1,2$\pm$1,2 & 10,8$\pm$5,0 & 1,7$\pm$1,1 & 6,4$\pm$2,8 & 4,2$\pm$1,9 \\
        APs in \% & \textbf{31,0$\pm$3,1} & 13,2$\pm$2,4 & 25,0$\pm$3,0 & 9,3$\pm$5,1 & 27,3$\pm$5,4 & 16,2$\pm$2,9 & 22,2$\pm$3,7 & 16,4$\pm$2,7 \\
        APm in \% & \textbf{31,7$\pm$4,3} & 16,3$\pm$2,9 & 22,5$\pm$3,3 & 14,0$\pm$3,8 & 27,5$\pm$4,6 & 18,6$\pm$3,2 & 22,4$\pm$3,0 & 20,3$\pm$3,2 \\
        AR in \% & \textbf{41,3$\pm$2,9} & 24,7$\pm$2,6 & 40,1$\pm$3,2 & 25,5$\pm$5,3 & 37,2$\pm$5,1 & 27,5$\pm$2,8 & 37,0$\pm$3,8 & 34,0$\pm$2,4 \\
        ARs in \% & 40,2$\pm$3,4 & 22,4$\pm$2,8 & \textbf{40,3$\pm$3,9} & 24,1$\pm$6,0 & 36,0$\pm$5,8 & 25,4$\pm$3,4 & 36,3$\pm$4,3 & 33,1$\pm$2,4 \\
        ARm in \% & \textbf{48,6$\pm$3,7} & 37,6$\pm$3,4 & 43,9$\pm$3,2 & 36,1$\pm$4,2 & 45,2$\pm$4,4 & 39,7$\pm$1,9 & 43,8$\pm$3,7 & 43,6$\pm$4,1 \\
        \bottomrule
    \end{tabular}
    \end{adjustbox}
\end{table}
\noindent
In \autoref{tab:ergebnisse-detr-2}, which summarizes the averaged results of \ac{detr} over the second dataset in terms of robustness, the pre-trained variants with the larger batch of 16 are shown to be the better ones. While the original weighting of the class loss produced the best results for the \ac{ap} metrics, the reduced weighting produced the best results in the \ac{ar} metrics. As already noted in \autoref{tab:ergebnisse-detr} for the results for the first testset, the values of the untrained combinations, which are equal or close to zero in all metrics, are also noticeable here. 

\begin{table}[h]
    \centering
    \caption[Results for the robustness of the DETR]{Results of various \ac{ap} and \ac{ar} metrics of the \ac{detr} on the second test dataset for robustness. The \acp{hpc} are given in the form batch size\_loss weighting\_training status, where \textbf{L} stands for the original and \textbf{l} for the 100-times reduced weighting, as well as \textbf{p} for pretrained and \textbf{u} for untrained.}
    \label{tab:ergebnisse-detr-2}
    \begin{adjustbox}{max width=\textwidth}
    \begin{tabular}{lcccccccc}
        \toprule
        \textbf{DETR} & \textbf{4\_L\_p} & \textbf{4\_L\_u} & \textbf{4\_l\_p} & \textbf{4\_l\_u} & \textbf{16\_L\_p} & \textbf{16\_L\_u} & \textbf{16\_l\_p} & \textbf{16\_l\_u} \\
        \midrule
        %AP & 0,286 & 0,000 & 0,026 & 0,000 & \textbf{0,307} & 0,000 & 0,249 & 0,000 \\
        %AP@50 & 0,714 & 0,000 & 0,056 & 0,000 & \textbf{0,756} & 0,000 & 0,611 & 0,000 \\
        %AP@75 & 0,162 & 0,000 & 0,021 & 0,000 & \textbf{0,178} & 0,000 & 0,147 & 0,000 \\
        %APs & 0,281 & 0,000 & 0,028 & 0,000 & \textbf{0,297} & 0,000 & 0,246 & 0,000 \\
        %APm & 0,343 & 0,000 & 0,030 & 0,000 & \textbf{0,388} & 0,000 & 0,320 & 0,000 \\
        %AR & 0,542 & 0,000 & 0,598 & 0,000 & 0,542 & 0,00004 & \textbf{0,620} & 0,000 \\
        %ARs & 0,533 & 0,000 & 0,587 & 0,000 & 0,531 & 0,00004 & \textbf{0,611} & 0,000 \\
        %ARm & 0,602 & 0,000 & 0,672 & 0,00008 & 0,611 & 0,000 & \textbf{0,688} & 0,000 \\
        AP in \% & 28,6$\pm$2,8 & 0,0$\pm$0,0 & 2,6$\pm$0,9 & 0,0$\pm$0,0 & \textbf{30,7$\pm$2,8} & 0,0$\pm$0,0 & 24,9$\pm$2,2 & 0,0$\pm$0,0 \\
        AP@50 in \% & 71,4$\pm$3,4 & 0,0$\pm$0,0 & 5,6$\pm$1,6 & 0,0$\pm$0,0 & \textbf{75,6$\pm$3,2} & 0,0$\pm$0,0   & 61,1$\pm$4,4 & 0,0$\pm$0,0  \\
        AP@75 in \% & 16,2$\pm$3,6 & 0,0$\pm$0,0 & 2,1$\pm$0,9 & 0,0$\pm$0,0 & \textbf{17,8$\pm$4,1} & 0,0$\pm$0,0   & 14,7$\pm$2,8 & 0,0$\pm$0,0 \\
        APs in \% & 28,1$\pm$3,1 & 0,0$\pm$0,0 & 2,8$\pm$0,9 & 0,0$\pm$0,0 & \textbf{29,7$\pm$2,9} & 0,0$\pm$0,0 & 24,6$\pm$2,3 & 0,0$\pm$0,0 \\
        APm in \% & 34,3$\pm$2,8 & 0,0$\pm$0,0 & 3,0$\pm$1,1 & 0,0$\pm$0,0 & \textbf{38,8$\pm$3,6} & 0,0$\pm$0,0 & 32,0$\pm$4,5 & 0,0$\pm$0,0 \\
        AR in \% & 54,2$\pm$2,9 & 0,0$\pm$0,0 & 59,8$\pm$3,2 & 0,0$\pm$0,0 & 54,2$\pm$2,3 & 0,004$\pm$0,02 & \textbf{62,0$\pm$2,2} & 0,0$\pm$0,0 \\
        ARs in \% & 53,3$\pm$3,4 & 0,0$\pm$0,0 & 58,7$\pm$4,0 & 0,0$\pm$0,0 & 53,1$\pm$2,5 & 0,004$\pm$0,02 & \textbf{61,1$\pm$2,6} & 0,0$\pm$0,0 \\
        ARm in \% & 60,2$\pm$2,6 & 0,0$\pm$0,0 & 67,2$\pm$2,6 & 0,008$\pm$0,04 & 61,1$\pm$3,2 & 0,0$\pm$0,0 & \textbf{68,8$\pm$2,8} & 0,0$\pm$0,0 \\
        \bottomrule
    \end{tabular}
    \end{adjustbox}
\end{table}

\begin{table}[h]
    \centering
    \caption[Results for robustness of DAB-DETR]{Results of various \ac{ap} and \ac{ar} metrics of \ac{dab-detr} on the second test dataset for robustness. The \acp{hpc} are given in the form batch-size\_loss-weighting\_training-status, where \textbf{L} stands for the original and \textbf{l} for the 100-times reduced weighting, as well as \textbf{p} for pretrained and \textbf{u} for untrained.}
    \label{tab:ergebnisse-dab-detr-2}
    \begin{adjustbox}{max width=\textwidth}
    \begin{tabular}{lcccccccc}
        \toprule
        \textbf{DAB-DETR} & \textbf{4\_L\_p} & \textbf{4\_L\_u} & \textbf{4\_l\_p} & \textbf{4\_l\_u} & \textbf{16\_L\_p} & \textbf{16\_L\_u} & \textbf{16\_l\_p} & \textbf{16\_l\_u} \\
        \midrule
        %AP & 0,323 & 0,005 & 0,240 & 0,000 & \textbf{0,328} & 0,003 & 0,258 & 0,000 \\
        %AP@50 & 0,790 & 0,027 & 0,583 & 0,000 & \textbf{0,792} & 0,015 & 0,670 & 0,000 \\
        %AP@75 & 0,183 & 0,001 & 0,145 & 0,000 & \textbf{0,194} & 0,000 & 0,129 & 0,000 \\
        %APs & 0,322 & 0,002 & 0,262 & 0,000 & \textbf{0,323} & 0,002 & 0,269 & 0,000 \\
        %APm & 0,375 & 0,021 & 0,222 & 0,000 & \textbf{0,406} & 0,009 & 0,298 & 0,000 \\
        %AR & 0,553 & 0,067 & \textbf{0,640} & 0,004 & 0,556 & 0,057 & 0,592 & 0,002 \\
        %ARs & 0,536 & 0,036 & \textbf{0,641} & 0,003 & 0,540 & 0,038 & 0,588 & 0,001 \\
        %Rm & 0,619 & 0,198 & \textbf{0,662} & 0,009 & 0,632 & 0,147 & 0,634 & 0,005 \\
        AP in \% & 32,3$\pm$2,0 & 0,5$\pm$0,2 & 24,0$\pm$4,4 & 0,0$\pm$0,0 & \textbf{32,8$\pm$3,3} & 0,3$\pm$0,2 & 25,8$\pm$3,4 & 0,0$\pm$0,0 \\
        AP@50 in \% & 79,0$\pm$2,0 & 2,7$\pm$1,0 & 58,3$\pm$8,4 & 0,0$\pm$0,0 & \textbf{79,2$\pm$2,9} & 1,5$\pm$1,0 & 67,0$\pm$4,8 & 0,0$\pm$0,0 \\
        AP@75 in \% & 18,3$\pm$3,1 & 0,1$\pm$0,2 & 14,5$\pm$4,7 & 0,0$\pm$0,0 & \textbf{19,4$\pm$5,7 }& 0,0$\pm$0,0 & 12,9$\pm$4,0 & 0,0$\pm$0,0 \\
        APs in \% & 32,2$\pm$2,3 & 0,2$\pm$0,2 & 26,2$\pm$4,7 & 0,0$\pm$0,0 & \textbf{32,3$\pm$3,4} & 0,2$\pm$0,2 & 26,9$\pm$3,8 & 0,0$\pm$0,0 \\
        APm in \% & 37,5$\pm$2,3 & 2,1$\pm$1,0 & 22,2$\pm$6,3 & 0,0$\pm$0,0 & \textbf{40,6$\pm$4,4} & 0,9$\pm$0,6 & 29,8$\pm$4,7 & 0,0$\pm$0,0 \\
        AR in \% & 55,3$\pm$2,4 & 6,7$\pm$2,1 & \textbf{64,0$\pm$3,8} & 0,4$\pm$0,5 & 55,6$\pm$3,0 & 5,7$\pm$3,3 & 59,2$\pm$2,4 & 0,2$\pm$0,3 \\
        ARs in \% & 53,6$\pm$2,8 & 3,6$\pm$2,0 & \textbf{64,1$\pm$4,3} & 0,3$\pm$0,5 & 54,0$\pm$3,6 & 3,8$\pm$2,9 & 58,8$\pm$2,8 & 0,1$\pm$0,3 \\
        ARm in \% & 61,9$\pm$2,3 & 19,8$\pm$5,9 & \textbf{66,2$\pm$3,2} & 0,9$\pm$1,4 & 63,2$\pm$2,8 & 14,7$\pm$7,3 & 63,4$\pm$2,8 & 0,5$\pm$0,9 \\
        \bottomrule
    \end{tabular}
    \end{adjustbox}
\end{table}
\noindent
The performance of \ac{dab-detr} on the second dataset is shown in \autoref{tab:ergebnisse-dab-detr-2}. Compared to \ac{detr} in the previous table, some values are higher. Also, as can be seen for the first dataset in \autoref{tab:ergebnisse-dab-detr}, the untrained combinations show strikingly low values across all metrics.  The pre-trained variant with the larger batch and original loss weighting has the best mean values in the \ac{ap} metrics. The corresponding variant with the smaller batch has the second-highest values. In the \ac{ar} metrics, the pre-trained variant with the smaller batch and the reduced loss achieved the highest values, followed by the corresponding variant with the larger batch.
\\
In general, when comparing the corresponding tables of the first and second dataset, it can be observed that most of the values of the second dataset are clearly below the values of the first dataset. Across all eight tables, none of the untrained combinations is leading in any metric. In addition, none of the untrained combinations with only a pre-trained backbone achieved a better average result for any of the metrics than the corresponding fully pre-trained combinations.  Especially for \ac{detr} and \ac{dab-detr}, the performance of the untrained \acp{hpc} was significantly lower or equal to zero.  Combinations with reduced weighting of the class loss often achieved better results in the metrics of \ac{ar}, while in the metrics of \ac{ap} the highest values always resulted from a combination with the original weighting.
\\
The following two figures visualize the results of the significance tests separately for one of the two datasets. The values used for this correspond to the mean values highlighted in bold from the previous tables as well as the 25 individual values that make up these mean values. According to the Shapiro-Wilk test, five of the results of the first dataset and two of the 32 results of the second dataset are not normally distributed. For the first dataset, these are the results of the \ac{ap} for the Faster \ac{r-cnn}, \ac{yolo}v8 and \ac{detr} models, as well as the \ac{ap}@50 of the \ac{dab-detr} and the \ac{ap}s for the Faster \ac{r-cnn}. For the second test dataset, it are the results of the \ac{ap}@75 of the \ac{yolo}v8 and the \ac{ar}s of the Faster \ac{r-cnn}. The test for significant differences between the models was therefore also carried out with the Kruskal-Wallis test in addition to the \ac{anova} for the metrics with non-normally distributed results. However, both methods yielded the same outcome: there are significant differences between the models in all metrics except \ac{ar} and \ac{ar}s of the first dataset. The subsequent tests to determine which models differed significantly from which other models were carried out with Dunn tests for comparisons with non-normally distributed result sets and with T-tests for all other comparisons.

\begin{figure}[h!]
    \centering
    \includegraphics[width=\linewidth,trim=2.5cm 0cm 3cm 1cm, clip]{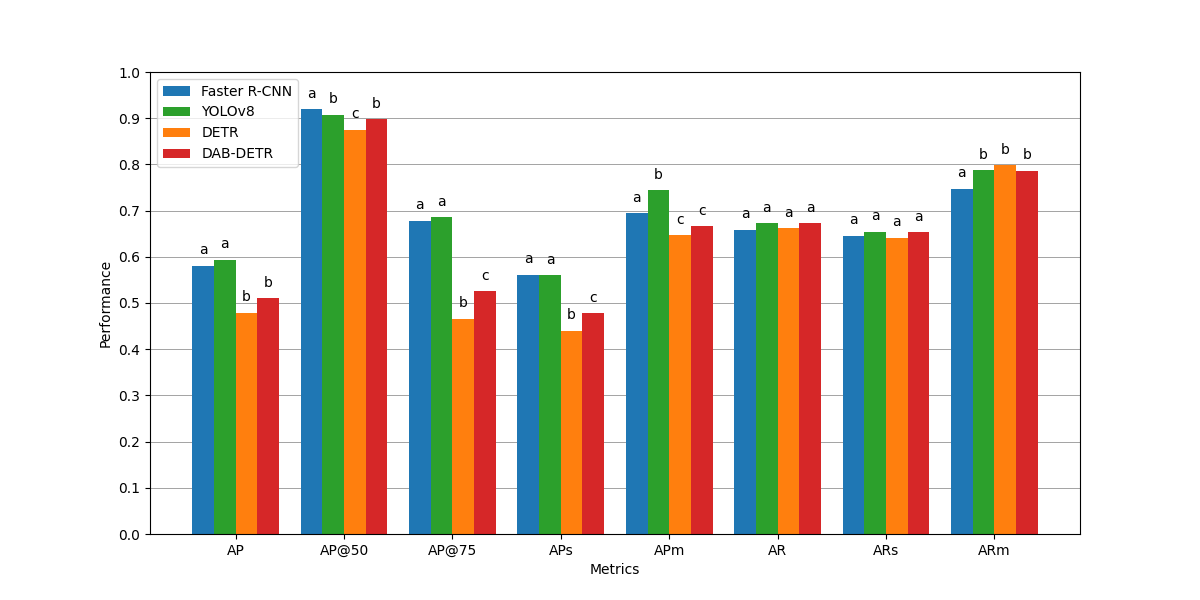}
    \caption[Significant differences on the first dataset]{Significant differences of the models on the first dataset in the respective metrics with the best \acp{hpc}. Different letters above the columns indicate statistically significant differences}.
    \label{fig:säule}
\end{figure}
\noindent
The mean values as well as the significance of the results determined with the Dunn and T-tests on the first dataset are shown in \autoref{fig:säule}. Models with the same letters within a metric have no significant difference to each other, while different letters show a statistically significant difference. Consequently, Faster \ac{r-cnn} and \ac{yolo}v8 are significantly better in the \ac{ap} than the two models with the transformer-based architecture. In the \ac{ap}@50, the Faster \ac{r-cnn} achieves the significantly highest value, followed by \ac{yolo}v8 and \ac{dab-detr}, between which there is no significant difference. \ac{detr} has the lowest significant value here. With a higher threshold of \ac{iou} of 75\% instead of 50\%, Faster \ac{r-cnn} and \ac{yolo}v8 are again significantly ahead of the transformers, whose performance drops significantly further. Here, \ac{detr} is again significantly below \ac{dab-detr}. The same ratios result for \ac{ap}s, that is over the small objects. There is no significant difference between the transformers for medium-sized objects. The Faster \ac{r-cnn} is significantly better than the transformers and \ac{yolo} achieves the significantly best performance. As already mentioned, there are almost no significant differences between the models in \ac{ar} on the first dataset. Only when considering the objects of medium size, the Faster \ac{r-cnn} is significantly below the other models. 

\begin{figure}[h!]
    \centering
    \includegraphics[width=\linewidth,trim=2.5cm 0cm 3cm 1cm, clip]{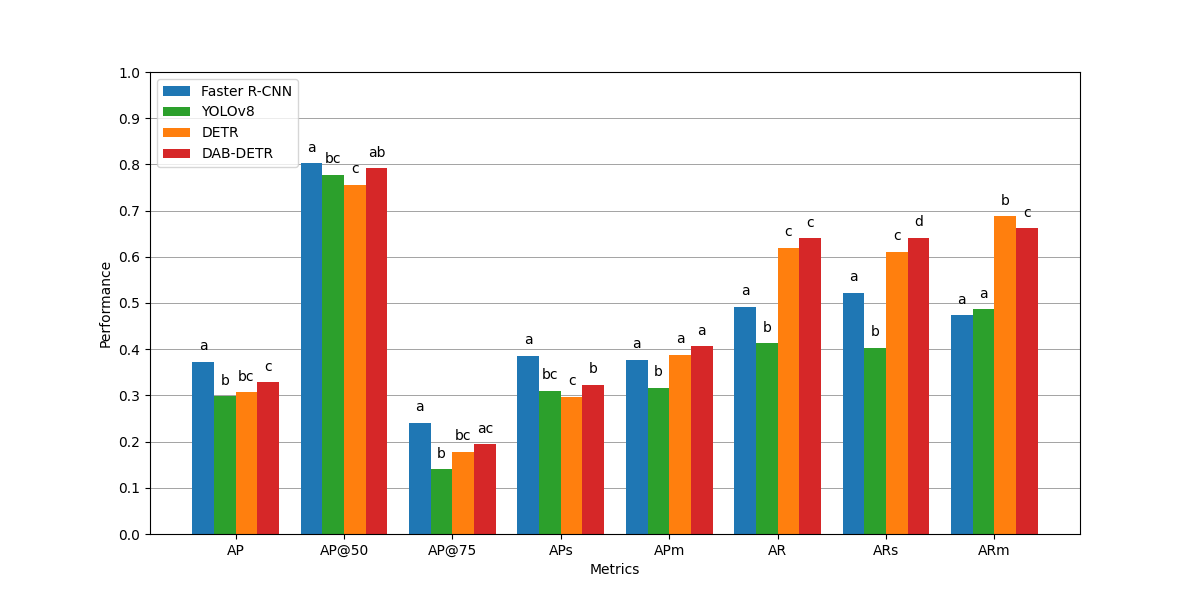}
    \caption[Significant differences on the second dataset]{Significant differences of the models on the second dataset in the respective metrics with the best \acp{hpc}. Different letters above the columns indicate statistically significant differences}.
    \label{fig:säule2}
\end{figure}
\noindent
\autoref{fig:säule2} illustrates the results of the significance tests for the second dataset, with which the robustness of the models was tested, according to the same principle as \autoref{fig:säule} shows the results of the first dataset. In the \ac{ap}, the Faster \ac{r-cnn} emerges as the significantly strongest model, followed by \ac{dab-detr}, which has a significantly higher performance than \ac{yolo}v8. \ac{detr} lies between \ac{dab-detr} and \ac{yolo}v8 and shows no significant difference to both models. In the \ac{ap}@50 and @75, the Faster \ac{r-cnn} together with the \ac{dab-detr} perform significantly better. However, the \ac{dab-detr} cannot be significantly separated from the \ac{yolo}v8 respectively from \ac{detr}. At the \ac{iou} of 50\% the \ac{detr} and at 75\% the \ac{yolo}v8 is the model with the lowest mean value and also significantly below \ac{dab-detr}. However, no difference can be detected between the two models in both metrics. Looking at the \ac{ap} only for the small objects, the Faster \ac{r-cnn} is significantly ahead of \ac{dab-detr}, which is significantly ahead of \ac{detr}. \ac{yolo}v8 is located between the two transformers without statistically detectable differences to them. If only the medium-sized objects are considered, \ac{yolo}v8 is the significantly worst model. The other three models, on the other hand, show no significant differences. For \ac{ar}, the results of the transformers are significantly higher and those of \ac{yolo}v8 significantly lower than those of the Faster \ac{r-cnn}. This also applies to \ac{ar}, which only relates to small objects. However, \ac{dab-detr} also differs significantly here as the better model compared to \ac{detr}. In the rightmost comparison of the diagram, which shows the performance of the models in \ac{ar} only for objects of medium size, Faster \ac{r-cnn} and \ac{yolo}v8 can be recognized without significant differences to each other with a significantly lower performance than the transformers. Between the transformers, \ac{detr} shows the significantly higher performance. 
\\
The comparison of both bar charts firstly illustrates that all models performed lower in all metrics on the second dataset. In addition, as already described in the context of the tables, \ac{yolo}v8 shows the largest differences between the results of the two datasets. This means that the Faster \ac{r-cnn} in the second dataset is significantly higher than the \ac{yolo}v8 in almost all metrics except \ac{ar}m. Furthermore, the performance of the transformers in the three \ac{ar} metrics drops significantly less from the first to the second dataset than the performance of the other two models. The smallest drop in performance from the first to the second dataset in the \ac{ap} metrics occurs when considering all object sizes with a \ac{iou} threshold of 50\%.

\section{Discussion}
\label{sec:Discussion}
\noindent
In this work, \ac{ai} models were trained and tested for the first time to detected animal excretions. The described and visualized results confirm the presumed general suitability of machine vision models for object detection of animal excretions in pig pens. Depending on the assumed exactness for the localization of the excretions in the image, slightly more than 90\% of the excretions declared by the models were also real excretions.  Furthermore, between 66\% and 67\% of all real excretions were correctly recognized. It should be noted here that the \ac{ar} was only averaged over the thresholds of the \ac{iou} in a range of 50\% to 95\%. If we compare the values with the values of the \ac{ap} over this range, the recall is higher than the precision. In addition, the \ac{iou} influences the recall more than the precision, as it converts \acp{tp} with inaccurate \acp{bbox} to \acp{fn}, which are both taken into account by the recall (see formula \eqref{eq:2}). Precision, on the other hand, does not take the \acp{fn} into account (see formula \eqref{eq:1}). If a \ac{tp} becomes a \ac{fn}, the precision decreases the numerator and denominator, while the recall decreases the numerator and the denominator remains constant. From this it can be concluded that the \ac{ar} for a lower \ac{iou} is probably at least as high as the \ac{ap} and 90\% or more of the excretions in the barn at a \ac{iou} of 50\% are also recognized as such. However, this can only be determined with certainty when such a metric is recorded in repeated tests.
\\
In addition, the results show significant differences between the four tested models in the observed metrics. Depending on the metric used, a different model is best for detecting the excretions. Across all object sizes and the different thresholds of \ac{iou}, the Faster \ac{r-cnn} and \ac{yolo}v8 have a significantly higher \ac{ap} than the transformers. The larger difference in the performance of \ac{detr} and \ac{dab-detr} between the \ac{ap}@50 and the \ac{ap}@75 indicates a less accurate localization of the objects in the image by the transformers. The differences between the four models are therefore smaller in the basic detection than in the detection of their position and size. In the \ac{ar}, a significantly worse performance of the Faster \ac{r-cnn} was found for the objects of medium size, but this does not seem to have an effect on the \ac{ar} for all object sizes. The same can be observed with the significantly better performance of the \ac{yolo}v8 in the \ac{ap} of the medium object sizes. The reason for this is probably the unequal distribution of small and medium-sized objects in the first dataset, which leads to a stronger weighting of the smaller objects in the metrics related to all objects. For the first dataset, the results suggest a better suitability of the Faster \ac{r-cnn} and the \ac{yolo}v8. The determined training and detection speeds are shortest for the \ac{yolo}v8. This is consistent with the expectations based on its architecture as a single-satge detector, while a lower accuracy compared to the Faster \ac{r-cnn} cannot be confirmed on the first dataset.
\\
Furthermore, important insights into the robustness of the models can be gained from the test results of the second dataset. In addition to the planned deviations in the angle and distance of the camera to the ground, as well as the different floor type, the second dataset also contains unexpected objects. These are primarily birds and ground surfaces heated by solar radiation. The objects in the second dataset therefore differ significantly in type, shape, structure and size. The results on the robustness of the models can therefore be considered valid and meaningful. The slight difference to the first dataset in the \ac{ap}@50 shows that up to 80\% of the excretions declared by the models are also real excretions on the second dataset and only the exact determination of the position decreases significantly. The transformers have the highest robustness in the \ac{ar} metrics and also recognize over 60\% of the existing excretions on the second dataset over the averaged range of \ac{iou} from 50\% to 95\%. The object sizes are more evenly distributed on the second dataset. They show a significantly better performance of the transformers for medium-sized objects in the \ac{ap} and a significantly better performance of \ac{detr} in the \ac{ar}m.  This corresponds to the advantage described in the literature for larger objects \citep{Software_DETR}. The \ac{yolo}v8 is the model with the lowest robustness in this work. The reason for this could lie in the architecture of the model, but also in the methodology of the work. On the one hand, the lower number of parameters of the model might not be sufficient to learn more complex but more general properties of the excretions, which the other models with more parameters were able to learn and to use on the second dataset. The generalization could therefore be limited by the lower number of parameters of the model. On the other hand, the high, unchanged number of epochs, together with the significantly earlier convergence of the metrics and the loss during training, indicates training beyond the epoch with the best generalization and thus an adaptation to the specific framework conditions of the first dataset. In other words, an overfitting of the model. In any case, further research is needed to determine the cause. A first starting point could be the use of early stopping in the configuration of \ac{yolo}v8.
\\
By varying the hyperparameters, the work provides an initial insight into the optimization potential of the models and the different effects of the same hyperparameters on different architectures. Adapting the batch size to the small dataset can therefore be a useful configuration. However, depending on the model and metric, \acp{hpc} with the original batch size also achieved the best performance. Reducing the weighting of the class loss did not lead to the expected general increase in the accuracy of \acp{bbox}. However, there was a significant advantage for recall at the expense of precision. Accordingly, all models tend to overestimate the excretions with the smaller loss weighting, which leads to an increase in \acp{tp} and \acp{fp}. The weighting of the loss thus turns out to be a hyperparameter that can be used specifically for setting precision and recall and should be investigated in more detail in future work. In terms of training status, the fully pre-trained models are clearly the better choice. An interesting observation is the conspicuous results of the untrained transformers. Despite a comparable number of trainable parameters with the Faster \ac{r-cnn} and the same pre-trained backbone, the transformers were not able to learn from the data to the same extent. This is confirmed by the very low values of \ac{detr} in \autoref{tab:epochen}, which correspond to the early convergence of the metrics and the loss observed during training. In contrast, the high values of the untrained \acp{hpc} with original loss weighting of \ac{dab-detr} in \autoref{tab:epochen} together with the non-converging loss suggest that the number of training epochs was too low here and that longer training could have led to higher performance of these \acp{hpc}.
\\
The comparison of the \acp{hpc} in the results of both datasets, as well as the comparison of the models on the second dataset, show the relevance of the weighting between precision and recall for the choice of models and their configuration. This is strongly related to the practical application in which the models are used. For the determination of the emission area, greater relevance could be placed on the recall in order to overestimate rather than underestimate the emissions. In conjunction with the automatic removal of excretions, on the other hand, precision could be of greater value, as a excretion that is detected but does not actually exist cannot be removed. A corresponding robot could otherwise be repeatedly occupied with the removal of non-existent excretions, while real excretions cause emissions during this time. If precision and recall are equally relevant, the F1 score could be used as a metric. An example for this could be an automatic treatment of emission areas with urease inhibitors, where all areas should be treated if possible, but the treatment of unnecessary areas should be avoided in order to save material and costs.  Consequently, it is not possible to make a generally valid statement about the more relevant metric, the best model or the better combination of hyperparameters.
\\
The creation and annotation of the datasets also resulted in an important knowledge gain. The datasets created are small for the training of \ac{ai} models for object detection and require little storage space due to the resolution of $640*480$ pixels and only one channel. Due to the special characteristics of the urine puddles and the data in the form of monochrome thermal images, it was not possible to use existing datasets with normal water puddles. However, as the results show, the information content of the small dataset was sufficient for training the models. Also it should be taken into account that only one class of objects was annotated or had to be learned by the models.  The quality of the images did not always allow a clear differentiation between feces and urine puddles. Since feces are at least indirectly relevant as a source of urease for the generation of emissions, the detection of feces is not a disadvantage anyway. The annotation of feces as a separate class would be advantageous since its area has a different effect on the formation of emissions than the area of urine puddles. For a clear separation in the annotation, higher resolutions or better image quality would be necessary. Another finding is that even if only images are required for training the models, information about the temperature progression over time and thus video recordings are still necessary for a correct and reliable annotation of the urine puddles on thermal images. 
\\
The methodology used for training the models deviates from the standard of nested cross-validation. But this is also primarily used for hyperparameter fitting, which was not the primary goal of this work. In order to pursue the objectives of the work in the best possible way, the deviations from nested cross-validation in the methodology were necessary. In addition, the deviations do not prevent a continuation of the nested cross-validation at a later point in time in the outer loop, as all training steps of the inner loop have been completed. It would therefore be possible to continue directly with the selection per inner loop and only the five selected hyperparameter combinations would have to be retrained in the outer loop. 
\\
As in any research work, there are also factors in this study that limit the validity of the results obtained. These include the assumptions made about the only scaling effects of the unvaried hyperparameters. The variation of the hyperparameters in this work only served to enable the models to be adapted to the specific dataset of the work. The assumptions were necessary in order to be able to examine the described comparisons between four different models in the context of one paper. However, it turned out that a deviation of the adopted hyperparameters partly led to better performance and that unmodified hyperparameters, such as the number of training epochs for \ac{yolo}v8 and \ac{dab-detr}, should be also adjusted in future research work. The best epochs from training were determined using \ac{ap}. This also means that the epochs are not automatically the best epochs for recall.  In most cases, the recall converged parallel to or before the precision during training. The differences between the best epochs in terms of recall or precision are therefore likely to be small. The number of n=25 results per hyperparameter combination and metric is somewhat low for the statistical significance tests. This therefore slightly limits the power of the significance test. However, most of the described significances are clearly below the significance levels and are therefore unambiguous. Lowering the significance level with the Bonferroni correction also ensures the results. Finally, the study is limited to pigs and their excretions. It is therefore not possible to transfer the results to other livestock species without further ado. Nevertheless, the basic suitability of \ac{ai} models for object detection of excretions should also be given for other livestock such as cattle by this work. At the same time, the limitations described show the need for further research and the potential for optimizing the models that is still available.

\section{Conclusion}
\label{sec:Conclusion}
\noindent
The obtained results as reported in this paper provide important novel insights into the field of automated excrement detection in livestock barns using \ac{ai}-based computer vision methods. They demonstrate the potential and capability of deep learning-based object detection models to recognize and locate urine puddles and feces in animal housing facilities, particularly for pig barns in this work. Significant differences between the four models tested are demonstrated. In particular, the demonstrated performance of the models with an accuracy of over 90\% is promising. It also provides an initial insight into the challenges to be overcome and the various strengths and weaknesses of the investigated models. Depending on the goal for which the detection is used, a different model or even a variant of the same model with different hyperparameters may be the better choice. This opens up new possibilities and methods for measuring and predicting emissions from livestock farming. It also shows that there is a need for further research, particularly with regard to the optimal setting of the hyperparameters. It was found that the hyperparameters have a significant influence on the performance of the models. Therefore, it is of great importance to conduct further research to find the optimal settings for the hyperparameters. In addition, there is also a need for research on the data required for urine puddle detection. Since various parameters such as housing type, animal species, animal age, sex and weather conditions can influence the characteristics of the puddles, it is important to collect and analyze suitable data to obtain a representative dataset and further improve the performance of the models. Overall, this work shows promising possibilities for the automated detection of urine puddles in livestock barns. The knowledge gained lays the basis for future research to optimize the performance and robustness of the models and to promote the use of this technology in the measurement, modeling and mitigation of emissions from livestock farming. As an additional outlook on further research, the extension of the object detection task, whose feasibility has been demonstrated in this work, to the computer vision task of pixel-wise segmentation provides another direction.

\section{Acknowledgements}
\label{sec:Acknowledgements}
\noindent
We thank the colleagues of Kiel University, especially Dr. Frauke Hagenkamp-Korth and the Institute of Agricultural Process Engineering of the Faculty of Agricultural and Nutritional Sciences for providing the raw thermal images and videos, which have served as the basis for creating the training data sets as created for this work.
\\\\
This work is financially supported by the Federal Ministry of Food and Agriculture (BMEL) based on a decision of the Parliament of the Federal Republic of Germany, granted by the Federal Office for Agriculture and Food (BLE); grant numbers 28N206507.
\\\\
The project is also supported by funds of the German Government's Special Purpose Fund held at Landwirtschaftliche
Rentenbank through the used video data collected in the EmiMin project.

\section{Declaration of generative AI and AI-assisted technologies in the writing process}
\label{sec:AI-Declaration}
\noindent
During the preparation of this work, the author(s) used DeepL (free version) in order to translate particular parts of the text and make linguistic optimizations. After using this tool/service, the author(s) thoroughly reviewed and edited the content as needed and take(s) full responsibility for the content of the publication.
\newpage
\printglossary[type=\acronymtype, title=Abbreviations] %Abkürzungen

\clearpage
\bibliographystyle{elsarticle-harv}
\bibliography{literature-bibtex_v2}



\newpage
\appendix
%%\input{Appendix/Appendix_1.tex}
%%\clearpage

\section{Additional information and figures regarding the deviations from the nested cross-validation in this work as described in \autoref{subsec:train_val_test}}
\label{appendix:B}
\begin{figure}[h!]
  \centering
  \includegraphics[width=0.9\linewidth]{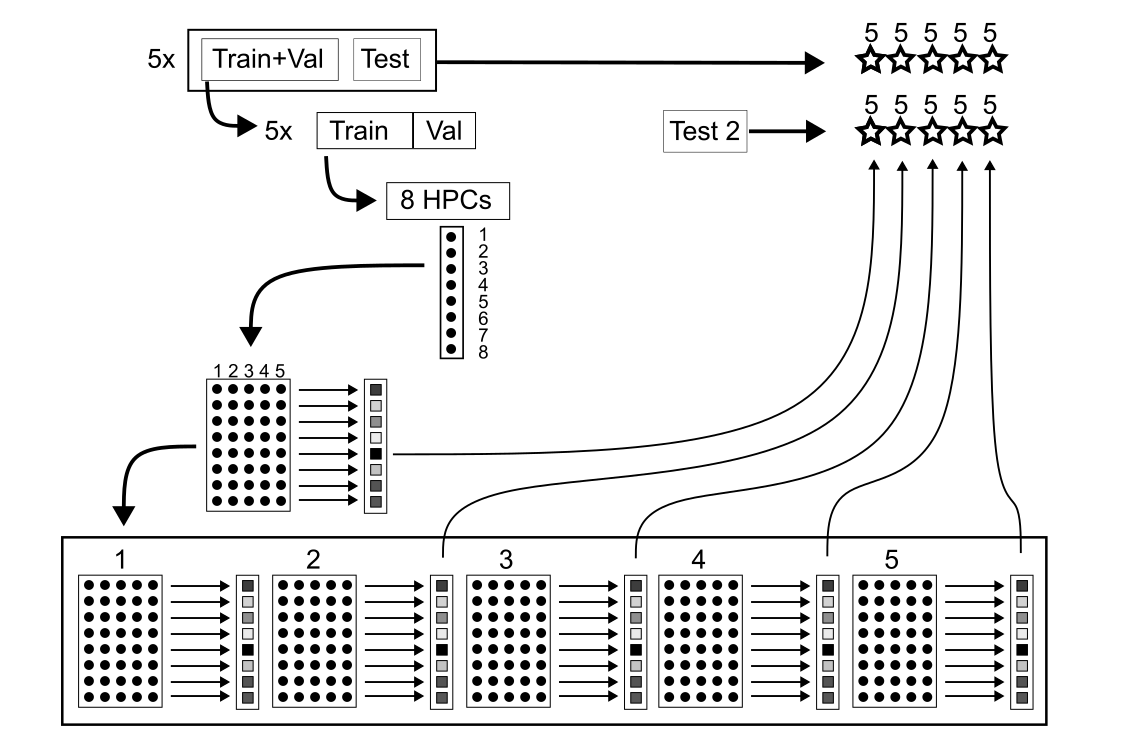}
  \caption[Schematic representation of a nested cross-validation]{Schematic representation of a nested cross-validation}.
  \label{fig:ncv1}
\end{figure}

\begin{figure}[h!]
  \centering
  \includegraphics[width=0.9\linewidth]{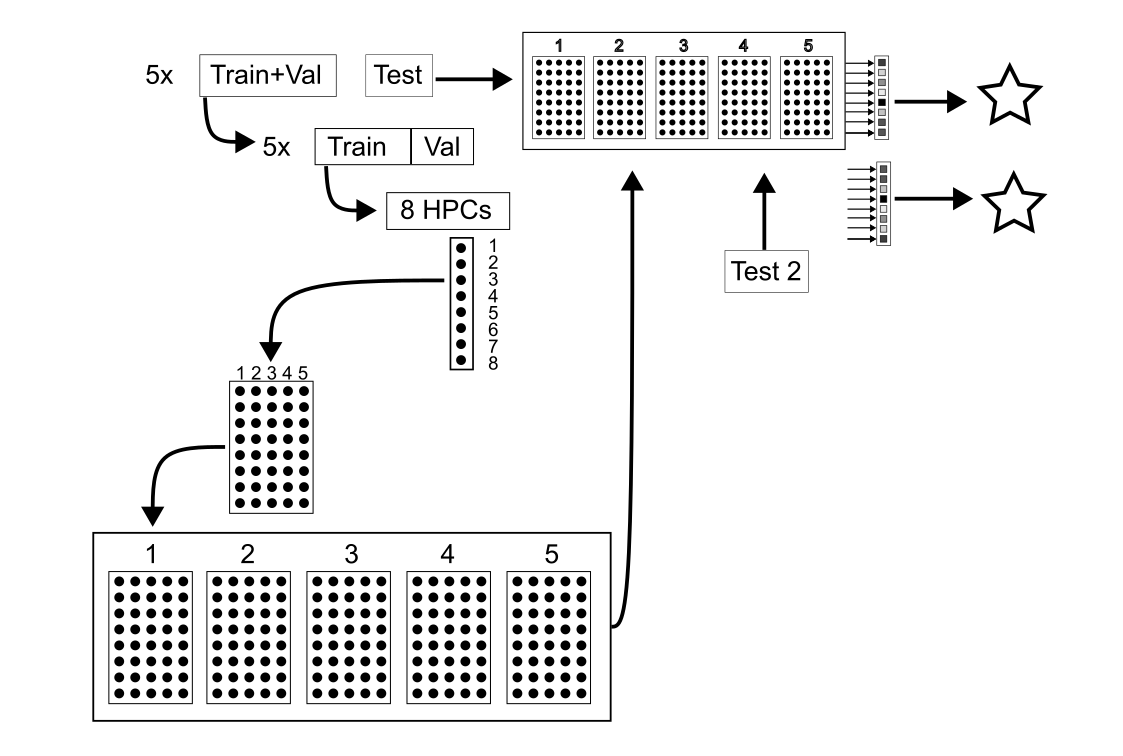}
  \caption[Schematic representation of the modified method]{Schematic representation of the method used in this work and modified from the nested cross-validation }.
  \label{fig:ncv2}
\end{figure}

\begin{figure}[h!]
  \centering
  \includegraphics[width=0.9\linewidth]{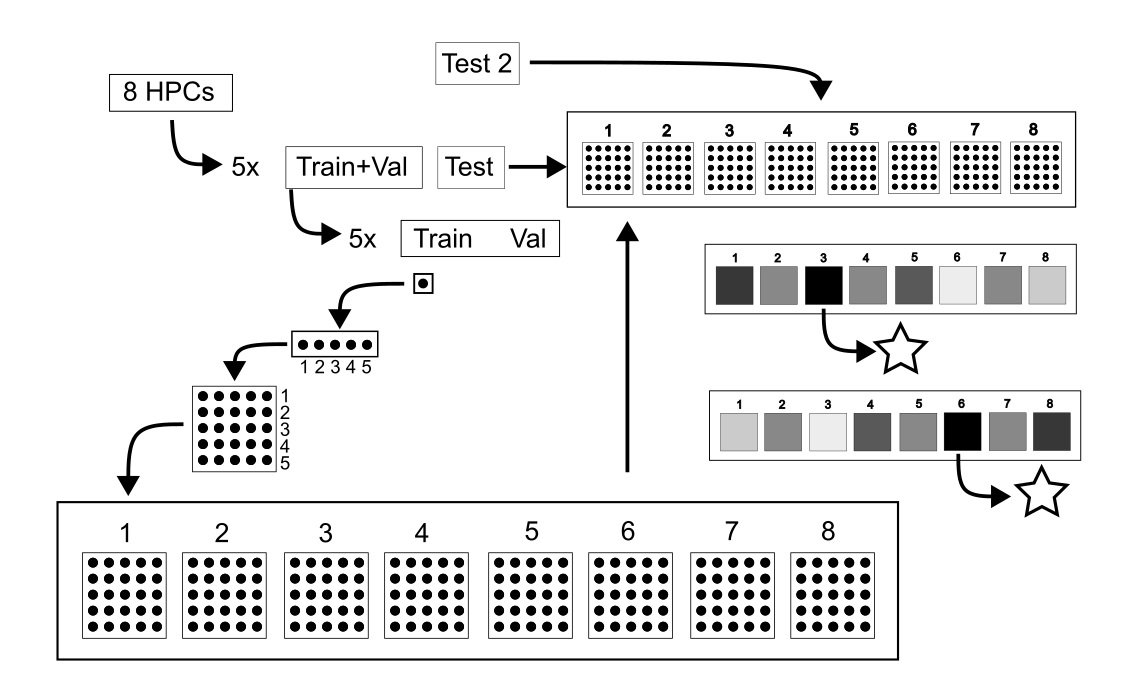}
  \caption[Schematic representation of the reshaped modified method]{Schematic representation of the reshaped modified method from \autoref{fig:ncv2}}
  \label{fig:ncv3}
\end{figure}

\noindent
The difference between the method used and nested cross-validation is that the selected models are not trained again, but tested directly and then selected. This allows each of the two test datasets to be applied to all \acp{hpc}. After all, an \ac{hpc} on the second test dataset that was not selected after the inner loop could be better for robustness than the selected combination. Another aspect is the possibility of selecting a different combination of hyperparameters as the best for each inner loop in nested cross-validation. In that case the result of the best combination would not be clear. With the modified method, the result is clear and with 25 instead of 5 values per \ac{hpc} also statistically more meaningful. The figures \ref{fig:ncv1}, \ref{fig:ncv2} and \ref{fig:ncv3} illustrate the nested cross-validation, the deviations that are made from it in this work and the resulting modified representation of the structure, in which the hyperparameters were moved from the inner loop to the outer loop. 

\clearpage
\section{Detailed description of the basic evaluation metrics required to compute average precision and average recall scores \autoref{subsec:evaluation}}
\label{appendix:C}
\noindent
There are a number of different metrics that make different but related statements. The metrics are usually calculated based on the results of the incorrectly or correctly detected objects or other metrics. A distinction is made between \acp{tp}, in this work a excretion that is also detected as a excretion, \acp{fp}, in this work a detected excretion that is not a excretion and \acp{fn}, in this work a excretion that is not detected as such. 

\subsection{Precision}
\noindent
The precision provides information about the proportion of real excretions in all excretions declared by the model. Formula \eqref{eq:1} is composed accordingly. It is therefore a question of how often a model incorrectly recognizes or declares a excretion. In other words, the more a model overestimates the number of excretions the lower the precision is. If the model only detected excretions that were actually excretions, then $\ac{fp} = 0$ and the precision would be 1. 

\begin{align}
    \label{eq:1}
    Precision{=} \frac{TP}{TP + FP} 
\end{align}

\subsection{Recall}
\noindent
The recall, on the other hand, provides a measure of the proportion of correctly detected excretions out of all excretions that should have been detected. Accordingly, \ac{tp} and \ac{fn} are included in the formula \eqref{eq:2}. It is therefore a question of how many of the real and annotated excretions are detected by the models. In other words, the more a model underestimates the number of excretions the lower the recall is. If the model detected all annotated excretions, then $\ac{fn} = 0$ and the recall would be 1.  

\begin{align}
    \label{eq:2}
    Recall{=} \frac{TP}{TP + FN} 
\end{align}

\subsection{F1 Score}
\noindent
As the precision does not take into account the proportion of \acp{fn} and the recall does not take into account the proportion of \acp{fp}, there is a risk that the other metric will become worse if only one of the two metrics is considered and optimized. The F1 score therefore combines both metrics and balances them. The formula for this is as shown in equation \eqref{eq:3}. Only if $\ac{fn} = 0$ and $\ac{fp} = 0$ and therefore $Precision = 1$ and $Recall = 1$, i.e. no excretion is detected too much or too little, would the $F1 score = 1$.

\begin{align}
    \label{eq:3}
    F1{=} 2 \cdot \frac{Precision \cdot Recall}{Precision + Recall} 
\end{align}

\subsection{Intersection over Union }
\begin{figure}[h]
  \centering
  \begin{subfigure}{0.49\linewidth}
    \centering
    \includegraphics[width=\linewidth]{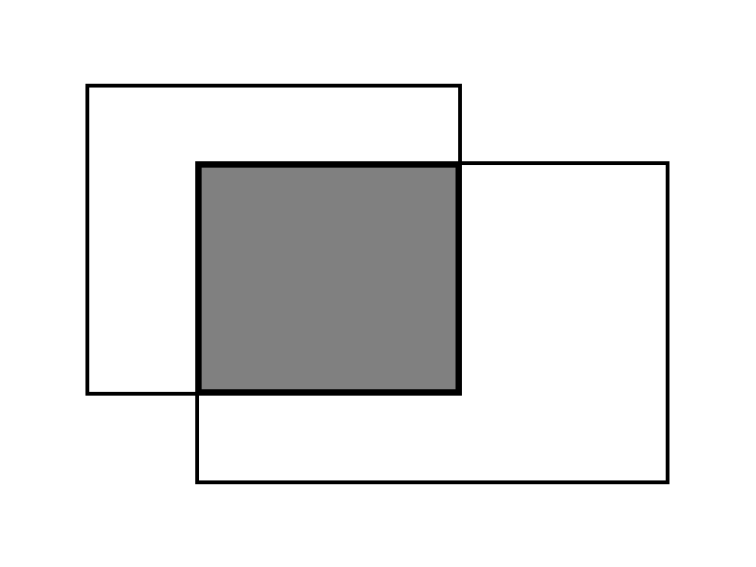}
    \caption{Intersection of two BBoxes.}
    \label{fig:Schnittmenge}
  \end{subfigure}
  \hfill
  \begin{subfigure}{0.49\linewidth}
    \centering
    \includegraphics[width=\linewidth]{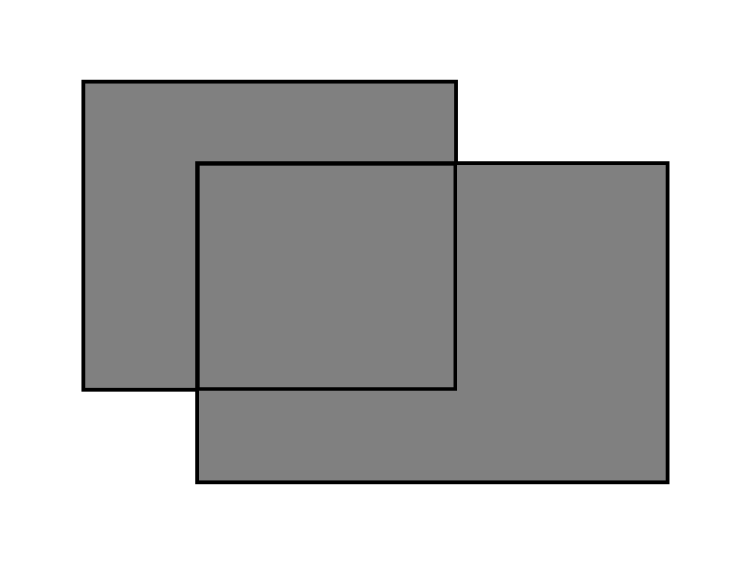}
    \caption{Union  of two BBoxes.}
    \label{fig:Vereingungsmenge}
  \end{subfigure}
  \caption{Parameters of the IoU.}
  \label{fig:iou}
\end{figure}

\noindent
While precision, recall and F1 score describe the basic detection and correct classification of the objects, the \ac{iou} deals with the size and loaclization of the \acp{bbox}.  A comparison is made between the \ac{bbox} predicted by the model and the \ac{bbox} from the \ac{gt} and the intersection and union of these \acp{bbox} are calculated with each other. This is illustrated in \autoref{fig:iou} and formula \eqref{eq:4}. If the value of \ac{iou} for a detected object is below a defined threshold value, then the object is evaluated as \ac{fn} instead of \ac{tp} even if the class is detected correctly.

\begin{align}
    \label{eq:4}
    IoU{=} \frac{Intersection}{Union} 
\end{align}

\end{document}